\documentclass[twoside,fleqn,espcrc2]{elsarticle}
\usepackage{lineno,hyperref}
\usepackage[hyphenbreaks]{breakurl}
\modulolinenumbers[5]

\usepackage{subfigure}
\usepackage{multirow}
\usepackage{bm}
\usepackage{verbatim}

\usepackage{color}

\usepackage[table,xcdraw]{xcolor}
\usepackage{booktabs}
\usepackage[numbers]{natbib}

\usepackage{graphicx} 
\usepackage{amssymb}
\usepackage{amsmath}
\usepackage{mathabx}
\usepackage{soul}

\usepackage{hhline}

\usepackage{epstopdf}

\usepackage{enumitem}
\usepackage[font=small,skip=2pt]{caption}
\graphicspath{ {Figures/}}

\newcolumntype{L}[1]{>{\raggedright\let\newline\\\arraybackslash\hspace{0pt}}m{#1}}
\newcolumntype{C}[1]{>{\centering\let\newline\\\arraybackslash\hspace{0pt}}m{#1}}
\newcolumntype{R}[1]{>{\raggedleft\let\newline\\\arraybackslash\hspace{0pt}}m{#1}}

\usepackage{ifthen}
\makeatletter
\renewcommand\@cite[2]{%
Ref.~#1\ifthenelse{\boolean{@tempswa}}
{, \nolinebreak[3] #2}{}
}
\renewcommand\@biblabel[1]{#1.}
\makeatother

\usepackage[textwidth=3.7cm]{todonotes}

\newcounter{cntcomment}

\newcommand{\CNNbase}{\mr{CNN}_\mr{base}}
\newcommand{\CNNsingle}{\mr{CNN}_\mr{single}}
\newcommand{\CNNmulti}{\mr{CNN}_\mr{multi}}

\newcommand{\mr}[1]{\mathrm{#1}}

\newcommand{\vold}[1]{$#1\!\times\!#1\!\times\!#1$}
\newcommand{\FreeSurfer}{\textit{FreeSurfer}}

\makeatletter
\def\ps@pprintTitle{%
 \let\@oddhead\@empty
 \let\@evenhead\@empty
 \def\@oddfoot{}%
 \let\@evenfoot\@oddfoot}
\makeatother

\begin{document}

\begin{frontmatter}

\title{3D fully convolutional networks for subcortical segmentation in MRI: A large-scale study }

\author[LIVIA]{Jose Dolz\corref{mycorrespondingauthor}}
\author[LIVIA]{Christian Desrosiers}
\author[LIVIA]{Ismail Ben Ayed}

\address[LIVIA]{LIVIA Laboratory, \'Ecole de technologie sup\'erieure (ETS), Montreal, QC, Canada}

\cortext[mycorrespondingauthor]{Corresponding author: jose.dolz.upv@gmail.com}
%
%
%

\begin{abstract}

This study investigates a 3D and fully convolutional neural network (CNN) for subcortical brain structure segmentation in MRI. 3D CNN architectures have been generally avoided due to their computational and memory requirements during inference. We address the problem via small kernels, allowing deeper architectures. We further model both local and global context by embedding intermediate-layer outputs in the final prediction, which encourages consistency between features extracted at different scales and embeds fine-grained information directly in the segmentation process. Our model is efficiently trained end-to-end on a graphics processing unit (GPU), in a single stage, exploiting the dense inference capabilities of fully CNNs. 

We performed comprehensive experiments over two publicly available datasets. First, we demonstrate a state-of-the-art performance on the ISBR dataset. Then, we report a {\em large-scale} multi-site evaluation over 1112 unregistered subject datasets acquired from 17 different sites (ABIDE dataset), with ages ranging from 7 to 64 years, showing that our method is robust to various acquisition protocols, demographics and clinical factors. Our method yielded segmentations that are highly consistent with a standard atlas-based approach, while running in a fraction of the time needed by atlas-based methods and avoiding registration/normalization steps. This makes it convenient for massive multi-site neuroanatomical imaging studies. To the best of our knowledge, our work is the first to study subcortical structure segmentation on such large-scale and heterogeneous data.

\end{abstract}


\end{frontmatter}


\section{Introduction}
\label{intro}

Accurate segmentation of subcortical brain structures is crucial to the study of various brain disorders such as schizophrenia \cite{van2016subcortical}, Parkinson \cite{geevarghese2015subcortical}, autism \cite{goldman2013motor} and multiple-sclerosis \cite{llado2012segmentation,garcia2013review}, as well as to the assessment of structural brain abnormalities \cite{koolschijn2009brain}. For instance, changes in the morphology and developmental trajectories of the caudate nucleus, putamen and nucleus accumbens have been associated with autism spectrum disorder (ASD), and may be linked to the occurrence of restricted and repetitive behaviors \cite{langen2009changes}. Accurate segmentation of these structures would help understanding such complex disorders, monitoring their progression and evaluating treatment outcomes. 

Automating subcortical structure segmentation remains challenging, despite the substantial research interest and efforts devoted to this computational problem. Clinicians still rely on manual delineations, a prohibitively time-consuming process, which depends on rater variability and is prone to inconsistency \cite{Deeley2011}. These issues impede the use of manual segmentation for very large datasets, such as those currently used in various multi-center neuroimaging studies. Therefore, there is a critical need for fast, accurate, reproducible, and fully automated methods for segmenting subcortical brain structures.

\subsection{Prior art}

A multitude of (semi-) automatic methods have been proposed for segmenting brain structures \cite{Dolz2015REV}. We can divide prior-art methods into four main categories: atlas-based methods \cite{lotjonen2010fast,wang2013multi}, statistical models \cite{babalola20083d,rao2008hierarchical}, deformable models \cite{yang20043d} and machine learning based classifiers \cite{powell2008registration,dolz2016supervised}. Atlas-based methods work by aligning one or several anatomical templates to the target image, via a linear or non-linear registration process, and then transferring segmentation labels from the templates to the image. Although these methods often provide satisfactory results, segmentation times are typically long (ranging from several minutes to hours) due to the complexity of registration steps. Furthermore, such methods may not be able to capture the full anatomical variability of target subjects (e.g., subjects of young age or with structural abnormalities), and can fail in cases of large misalignments or deformations. Unlike atlas-based methods, approaches based on statistical models use training data to learn a parametric model describing the variability of specific brain structures (e.g., shapes, textures, etc.). When the number of training images is small compared to the number of parameters to learn, these approaches might result in overfitting the data, thereby introducing bias in the results. The robustness of such statistical approaches might also be affected by the presence of noise in training data. Finally, because parameters are updated iteratively by searching in the vicinity of the current solution, an accurate initialization is required for such approaches to converge to the correct structure. Unlike statistical models, segmentation techniques using deformable models do not require training data, nor prior knowledge. Because they can evolve to fit any target structure, such models are considered to be highly flexible compared to other segmentation methods. Yet, deformable models are quite sensitive to the initialization of the segmentation contour and the stopping criteria, both of which depend on the characteristics of the problem. The last category of methods, based on machine learning, uses training images to learn a predictive model that assigns class probabilities to each pixel/voxel. These probabilities are sometimes used as unary potentials in standard regularization techniques such as graph cuts \cite{shakeri2016sub}. Recently, machine learning approaches have achieved state-of-the-art performances in segmenting brain structures \cite{Dolz2015REV,powell2008registration}. Nevertheless, these approaches usually involve heavy algorithm design, with carefully engineered, application-dependent features and meta-parameters, which limit their applicability to different brain structures and modalities.

Deep learning has recently emerged as a powerful tool, achieving state-of-the art results in numerous applications of pattern or speech recognition. Unlike traditional methods that use hand-crafted features, deep learning techniques have the ability to learn hierarchical features representing different levels of abstraction, in a data-driven manner. Among the different types of deep learning approaches, convolutional neural networks (CNNs) \cite{lecun1998gradient,krizhevsky2012imagenet} have shown outstanding potential for solving computer vision and image analysis problems. Networks of this type are typically made up of multiple convolution, pooling and fully-connected layers, the parameters of which are learned using back-propagation. Their advantage over traditional architectures come from two properties: local-connectivity and parameter sharing. Unlike in typical neural nets, units in hidden layers of a CNN are only connected to a small number of units, corresponding to a spatially localized region. This reduces the number of parameters in the net, which limits memory/computational requirements and reduces the risk of overfitting. Moreover, CNNs also reduce the number of learned parameters by sharing the same basis function (i.e., convolution filters) across different image locations.

In biomedical imaging, CNNs have been recently investigated for several neuroimaging applications \cite{ciresan2012deep,zhang2015deep,havaei2016brain,pereira2016brain}. For instance, Ciresan et al. \cite{ciresan2012deep} used a CNN to accurately segment neuronal membranes in electron microscopy images. In this study, a sliding-window strategy was applied to predict the class probabilities of each pixel, using patches centered at the pixels as input to the network. An important drawback of this strategy is that its label prediction is based on very localized information. Moreover, since the prediction must be carried out for each pixel, this strategy is typically slow. 
Zhang et al. \cite{zhang2015deep} presented a CNN method to segment three brain tissues (white matter, gray matter and cerebrospinal fluid) from multi-sequence magnetic resonance imaging (MRI) images of infants. As inputs to the network, 2D images corresponding to a single plane were used. Deep CNNs were also investigated for glioblastoma tumor segmentation \cite{havaei2016brain}, using an architecture with several pathways, which modeled both local and global-context features. Pereira et al. \cite{pereira2016brain} presented a different CNN architecture for segmenting brain tumors in MRI data, exploring the use of small convolution kernels. Closer to this work, several recent studies investigated CNNs for segmenting subcortical brain structures \cite{shakeri2016sub,lee2011towards,moeskops2016automatic,milletari2016hough,Brebisson2015deep}. For instance, Lee et al. \cite{lee2011towards} presented a CNN-based approach to learn discriminative features from expert-labelled MR images. The study in \cite{moeskops2016automatic} used CNNs to segment brain structures in images from five different datasets, and reported performance for subjects in various age groups (ranging from pre-term infants to older adults). A multiscale patch-based strategy was used to improve these results, where patches of different sizes were extracted around each pixel as input to the network. 

Although medical images are often in the form of 3D volumes (e.g., MRI or computed tomography scans), most of the existing CNN approaches use a slice-by-slice analysis of 2D images. An obvious advantage of a 2D approach, compared to one using 3D images, is its lower computational and memory requirements. Furthermore, 2D inputs accommodate using pre-trained nets, either directly or via transfer learning. However, an important drawback of such an approach is that anatomic context in directions orthogonal to the 2D plane is completely discarded. As discussed recently in \cite{milletari2016hough}, considering 3D MRI data directly, instead of slice-by-slice, can improve the performance of a segmentation method. To incorporate 3D contextual information, de Brebisson et al. used 2D CNNs on images from the three orthogonal planes \cite{Brebisson2015deep}. The memory requirements of fully 3D networks were avoided by extracting large 2D patches from multiple image scales, and combining them with small single-scale 3D patches. All patches were assembled into eight parallel network pathways to achieve a high-quality segmentation of 134 brain regions from whole brain MRI. More recently, Shakeri et al. \cite{shakeri2016sub} proposed a CNN scheme based on 2D convolutions to segment a set of subcortical brain structures. In their work, the segmentation of the whole volume was first achieved by processing each 2D slice independently. Then, to impose volumetric homogeneity, they constructed a 3D conditional random field (CRF) using scores from the CNN as unary potentials in a multi-label energy minimization problem.

So far, 3D CNNs have been largely avoided due to the computational and memory requirements of running 3D convolutions during inference. However, the ability to fully exploit dense inference is an important advantage of 3D CNNs over 2D representations \cite{szegedy2015going}. While standard CNN approaches predict the class probabilities of each pixel independently from its local patch, fully convolutional networks (FCNNs) \cite{long2015fully} consider the network as a large non-linear filter whose output yields class probabilities. This accommodates images of arbitrary size, as in regular convolution filters, and provides much greater efficiency by avoiding redundant convolutions/pooling operations. Recently, 3D FCNNs yielded outstanding segmentation performances in the context of brain lesions \cite{brosch2015deep,kamnitsas2016efficient}.

\subsection{Contributions}

This study investigates a 3D and fully convolutional neural network for subcortical brain structure segmentation in MRI. Architectures using 3D convolutions have been generally avoided due to their computational and memory requirements during inference and, to the best of our knowledge, this work is the first to examine 3D FCNNs for subcortical structure segmentation. We address the problem via small kernels, allowing deeper architectures. We further model both local and global context by embedding intermediate-layer outputs in the final prediction, which encourages consistency between features extracted at different scales and embeds fine-grained information directly in the segmentation process. This contrasts with previous architectures (e.g., \cite{kamnitsas2016efficient}), where global context is modelled using separate pathways and low-resolution images. Our model is efficiently trained end-to-end on a graphics processing unit (GPU), in a single learning stage, exploiting the dense inference capabilities of FCNNs. Compared to conventional approaches, which typically require time-con\-su\-ming and error-prone registration steps, the proposed method also has the advantage of being alignment independent. This property is of great importance in clinical applications where scans from different subjects, modalities and acquisition protocols need to be analyzed.

We performed comprehensive experiments over two publicly available datasets. The IBSR dataset is first used to compare our method to existing approaches for subcortical brain segmentation, and demonstrate its state-of-the-art performance. We then report a large-scale evaluation over 1112 unregistered subject data from the multi-site ABIDE dataset, with ages ranging from 7 to 64 years, showing that our method is robust to various acquisition protocols, demographics and clinical factors. Our method yielded segmentations that are highly consistent with a standard atlas-based approach, while running in a fraction of the time needed by such methods.

This makes it convenient for massive multi-site neuroanatomical imaging studies. We believe our work is the first to assess subcortical structure segmentation on such large-scale and heterogeneous data.

\section{Methods and materials}
\label{sec:methods}

We start by presenting the proposed 3D FCNN architecture, which is at the core of our segmentation method. Sections \ref{ssec:smallKernels} and \ref{ssec:combiningFeatures} then describe how this architecture can be improved by additional convolution layers with smaller kernels, and by considering multiscale information from intermediate convolutional layers. Thereafter, Section \ref{ssec:preproc} presents the pre- and post-processing steps performed by our method on the data and output segmentations. Finally, Section \ref{ssec:experiments} focuses on the study design and experimental setup, providing information on the datasets used in the study, implementation details of the tested network architectures, and the metrics used to evaluate the performance of these architectures.  

\subsection{The proposed 3D FCNN architecture}
\label{ssec:3DFullyCNN}

\begin{figure}[h!]
     \begin{center} 

        \includegraphics[width=0.995\textwidth]{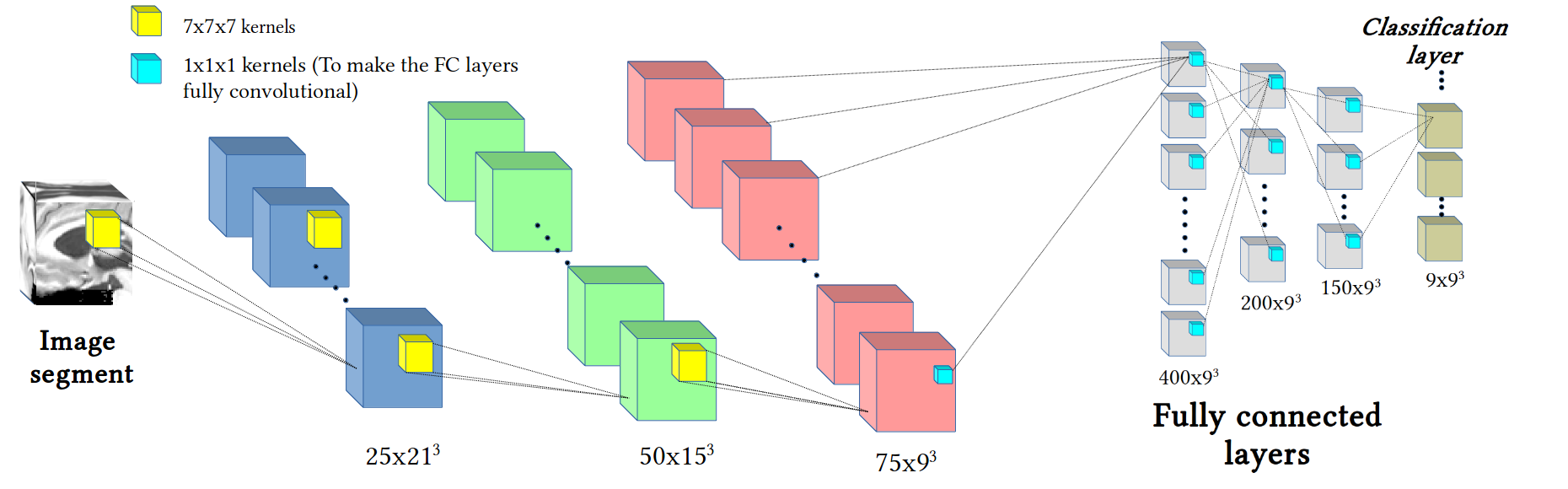}
\caption{The baseline FCNN architecture ($\CNNbase$), composed of 3 convolutional layers with kernels of size \vold{7}. This FCNN is obtained from a standard CNN by replacing the fully connected layers by a set of \vold{1} convolutional filters.}
        \label{fig:CNNBase}
\end{center}        
\end{figure}

Traditional CNN architectures, like AlexNet or GoogLeNet, require an input image of fixed size and use pooling layers to gradually reduce the spatial resolution of the representation. Spatial information is then completely discarded in fully-connected layers at the end of the network. Although originally designed for image recognition and classification tasks, such architectures can be used for semantic segmentation via a sliding-window approach, where regions defined by the window are processed one-by-one. However, this strategy has a low efficiency, due to the many redundant convolution and pooling operations. Processing regions in the image independently, instead of as a whole, also reduces segmentation accuracy. Fully Convolutional Networks (FCNNs) mitigate these limitations by treating the network as a single non-linear convolution, trained end-to-end \citep{long2015fully}. Unlike traditional CNNs, FCNNs are only composed of convolutional layers, allowing them to be applied to images of arbitrary size. Moreover, because the spatial map of class scores is obtained in a single dense inference step, FCNNs can avoid redundant convolution and pooling operations, which makes them computationally more efficient. 

The proposed segmentation method is based on a 3D fully CNN (FCNN) architecture, depicted in Figure \ref{fig:CNNBase}. This architecture is composed of three convolutional layers, each one containing several 3D convolution filters (or \emph{kernels}). Each filter in a layer is applied to the output of the previous layer, or the input volume in the case of the first layer, and the result of this operation is known as a feature map. Denote as $m_l$ the number of convolution kernels in layer $l$ of the network, and let $x^n_{l-1}$ be the 3D array corresponding to the $n$-th input of layer $l$. The $k$-th output feature map of layer $l$ is then given by:
\begin{equation}
\qquad \qquad \qquad \qquad 
    y^k_l \ = \ f\Big(\sum^{m_{l-1}}_{n=1} W^{k,n}_{i} \otimes x^n_{l-1} + b^{k}_{l}\Big),
\end{equation} 
where $W^{k,n}_{i}$ is a filter convolved (represented by $\otimes$) with each of the previous layers, $b^{k}_{l}$ is the bias, and $f$ is a non-linear activation function. Note that feature maps produced by convolutions are slightly smaller than their input volumes, the size difference along each dimension equal to the filter size in this dimension, minus one voxel. Hence, applying a \vold{3} convolution filter will reduce the input volume by 2 voxels along each dimension. A stride may also be defined for each convolutional layer, representing the displacement of the filter, along the three dimensions, after each application. 

In some FCNN architectures, pooling layers may be added between convolutional layers to reduce spatial resolution and, thus, the number of parameters to learn. Such layers can be interpreted as simple convolutional layers with non-unit stride (e.g., a stride of 2). The resolution of the input image is recovered by adding deconvolution (or \emph{transpose} convolution) layers at the end of the network \cite{long2015fully}. However, this strategy may lead to coarse segmentations. In the proposed architecture, we preserve spatial resolution by avoiding pooling layers and using a unit stride for all convolutional layers.

For the activation function, we used the Parametric Rectified Linear Unit (PReLU) \cite{he2015delving} instead of the popular Rectified Linear Unit (ReLU). This function can be formulated as
\begin{equation}
	\qquad \qquad \qquad \qquad	f(x_i) \ = \ \max(0,x_i) \, + \, a_i \! \cdot \! \min(0,x_i),
\end{equation}
where $x_i$ defines the input signal, $f(x_i)$ represents the output, and $a_i$ is a scaling coefficient for when $x_i$ is negative. While ReLU employs predefined values for $a_i$ (typically equal to 0), PReLU requires learning this coefficient. Thus, this activation function can adapt the rectifiers to their inputs, improving the network's accuracy at a negligible extra computational cost. 

As in standard CNNs, fully-connected layers are added at the end of the network to encode semantic information. However, to ensure that the network contains only convolutional layers, we use the strategy described in \cite{long2015fully} and \cite{kamnitsas2016efficient}, in which fully-connected layers are converted to a large set of \vold{1} convolutions. Doing this allows the network to retain spatial information and learn the parameters of these layers as in other convolutional layers. Lastly, neurons in the last layer (i.e., the classification layer) are grouped into $m=C$ feature maps, where $C$ denotes the number of classes. The output of the classification layer $L$ is then converted into normalized probability values via a softmax function. The probability score of class $c \in \{1, \ldots, C\}$ is computed as follows:
\begin{equation}
    \qquad \qquad \qquad \qquad \
        p_c \ = \ \frac{\exp\big(y^c_L\big)}{\sum^{C}_{c'=1} \exp\big(y^{c'}_L\big)}.
\end{equation}

The 3D FCNN architecture described in this section constitutes our baseline model for segmentation. In the following two sections, we describe how a deeper architecture can be achieved with smaller convolution kernels and how multiscale information can be added by combining features from intermediate convolutional layers.

\subsection{Deeper architecture via small convolution kernels} 
\label{ssec:smallKernels}

\begin{figure}[h!]
     \begin{center} 

        \includegraphics[width=0.995\textwidth]{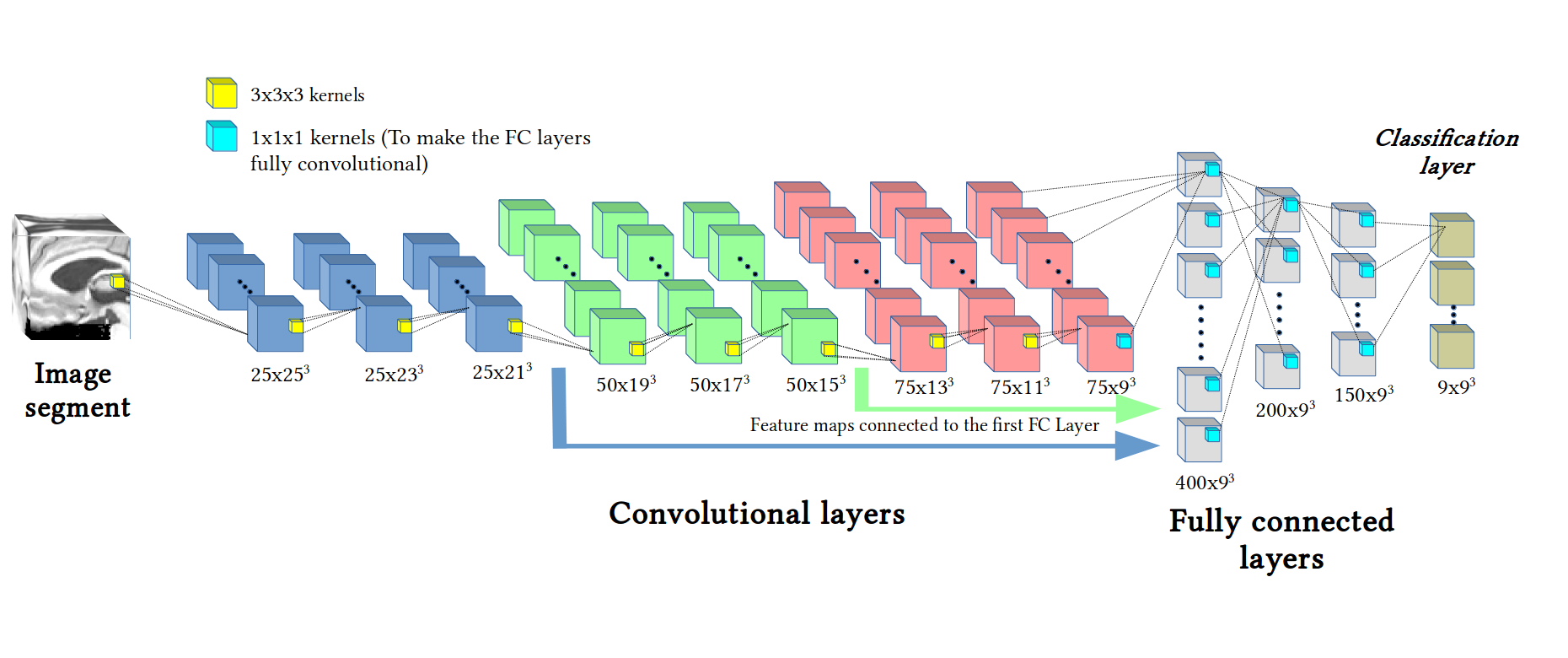}
\caption{The proposed improvements to the baseline architecture $\CNNbase$. A deeper architecture is achieved via smaller convolution kernels (\vold{3}). Also, multiscale information is used in the segmentation by injecting feature maps of intermediate layers (convolutional layers 3, 6 and 9) directly in the fully-connected layers, as represented by the arrows. We refer as $\CNNsingle$ the architecture with small kernel but a single scale, and as $\CNNmulti$ the architecture including both small kernels and multiscale information.}
 \label{fig:CNN_archit}
\end{center}        
\end{figure}

Numerous studies have shown the benefits of using deeper network architectures. In FCNNs, however, having many convolutions layers reduces the resolution of feature maps at the end of the network, leading to a coarse segmentation. To alleviate this problem, we extend the architecture of Figure \ref{fig:CNNBase} by replacing each convolutional layer by three successive convolutional layers with the same number of kernels, but smaller kernel sizes: \vold{3} instead of \vold{7}. The resulting topology is shown in Figure \ref{fig:CNN_archit}.

By using these smaller kernels, we obtain a deeper architecture while having fewer parameters in the network. Consequently, the network can learn a more complex hierarchy of features, with a reduced risk of overfitting. This fact is supported by the findings reported in \cite{simonyan2014very} for 2D CNNs, and in \cite{kamnitsas2016efficient} for 3D CNNs. 

\subsection{Combining features across multiple scales} 
\label{ssec:combiningFeatures}

In CNNs, the sequence of layers encodes features representing increasing levels of abstraction: the first convolutional layer typically models simple edge or blob detectors, whereas convolutional layers directly before the fully-connected ones model larger-scale and more complex structures. In various computer vision problems, like optical flow \cite{brox2004high}, reasoning across multiple levels of abstraction (i.e., \emph{scales}) has proven beneficial. Following this principle, we further improve our baseline FCNN to include multiscale information in the segmentation.

The multiscale version of our segmentation model, shown in Figure \ref{fig:CNN_archit}, has a single 3D image as input (highest available resolution) and combines the feature maps of various intermediate layers (i.e., convolutional layers 3, 6 and 9) in the fully-connected layers. This strategy \cite{farabet2013learning,chen2014semantic,chen2015semantic,hariharan2015hypercolumns} has two important advantages over other multiscale approaches, where the input image is resampled at multiple scales of resolution before being fed to the network \cite{kamnitsas2016efficient}. First, because it has a single set convolution filters at each layer, instead of one per pathway, the features at different scales are more likely to be consistent with each other. Moreover, since features from intermediate layers are injected in top-most layers, fine-grained information is used directly in the segmentation process. However, a drawback of this strategy is that it introduces a large number of parameters in fully-connected layers, which can make learning these parameters computationally complex.

Due to repeated convolution operations, the feature maps that need to be combined at the end of the network have different sizes: \vold{21} in layer 3, \vold{15} in layer 6, and \vold{9} in layer 9. Instead of vectorizing these features maps and appending them to one another, we extract \vold{9} volumes around the center of these feature maps, which are then stacked on top of each other. These volumes, which are compatible in size, encode different resolutions and receptive fields of the input image (i.e., the volume of layer 3 has a smaller receptive field but a higher resolution than the volumes of layers 6 and 9). Note that this technique leads to less parameters than vectorizing the full feature maps.

\subsection{Pre- and post-processing steps}
\label{ssec:preproc}

Data pre-processing steps are often required to ensure the performance of segmentation methods. Typical pre-processing steps for MRI data include the removal of non-brain tissues like the skull, as well as bias field correction. For multi-subject or longitudinal studies, additional steps are often necessary to normalize intensities or align volumes across multiple scans. In \cite{shakeri2016sub}, Shakeri et al. used registered and normalized MRI volumes to validate their subcortical parcellation method. Such elaborate data pre-processing has several disadvantages. First, aligning volumes to a template (e.g., MNI space) is a time-consuming operation, which would remove the computational benefit of using CNNs over atlas-based methods. Furthermore, training the network using data with a very specific and strict pre-processing reduces the network's ability of generalizing to unprocessed data, or data pre-processed differently.

In order to make our method robust to different imaging protocols and parameters, we used a simple pre-processing step that includes volume-wise intensity normalization, bias field correction and skull-stripping. The first two transforms, both computationally inexpensive, are used to reduce the sensitivity of the network to contrast and intensity bias. Skull-stripping, although more time-consuming, can be performed without registration (e.g., see \cite{smith2002fast}). This step is used mostly to reduce the size of the input image by discarding non-interesting areas, and thus unnecessary computations.   

Although the segmentations obtained using our network are generally smooth and close to manual labels, small isolated regions can sometimes appear in the segmentation. As post-processing step, we remove these small regions by keeping only the largest connected component from each class. Note that standard regularization approaches like CRFs \cite{lafferty2001conditional} have also been tested, but did not lead to significant improvements in accuracy. 

\subsection{Study design and experiment setup}
\label{ssec:experiments}

\subsubsection{Datasets}
\label{sssec:dataset}

The proposed segmentation method was tested on the following two publicly available datasets.

\paragraph{\textbf{IBSR}}

A set of 18 T1-weighted MRI scans from the Internet Brain Segmentation Repository (IBSR) was employed to obtain quantitative measures of performance and compare our method against competing approaches. These images were acquired at the Massachusetts General Hospital and are freely available at \url{http://www.cma.mgh.harvard.edu/ibsr/data.html}. In addition, the dataset also contains expert-labelled segmentations of 45 brain structures. Among these, a subset of 8 important subcortical structures were considered in this work: left and right thalamus, caudate, putamen, and pallidum. These structures were used in recent studies on brain parcellation (e.g., see \cite{shakeri2016sub}). All volumes have a size of $256\!\times\!256\!\times\!128$ voxels, with voxel sizes ranging from $0.8\!\times\!0.8\!\times\!1.5 \textrm{ mm}^3$ to $1.0 \!\times\!1.0\!\times\!1.5 \textrm{ mm}^3$.  To get unbiased estimates of performance, and following the validation methodolody of \cite{shakeri2016sub}, we employed a 6-fold cross validation strategy, where each fold is composed of 12 training examples (i.e., subjects), 3 validation examples and 3 test examples.

\paragraph{\textbf{ABIDE}}

The Autism Brain Imaging Data Exchange (ABIDE) \cite{di2014autism} was used as a second dataset in our experiments. ABIDE I involved 17 international sites, sharing previously collected resting state functional magnetic resonance imaging (R-fMRI), anatomical and phenotypic datasets made available for data sharing with the broader scientific community. This effort yielded a huge dataset containing 1112 subjects, including 539 from individuals with autism spectrum disorder (ASD) and 573 from typical controls (ages 7-64 years, median 14.7 years across groups). Characteristics for each site are presented in Table \ref{tab:sites}. 

\begin{table}[htb!]
\centering
\footnotesize
\begin{tabular} {lcccc}
\toprule
\addlinespace
\parbox{4.8cm}{\textbf{Site}} & 
\parbox{1.cm}{\textbf{Number \\ images}} &
\parbox{1.7cm}{\textbf{Voxel size \\ (mm$^3$) } }  &
\parbox{1.3cm}{\textbf{Control \\vs. ASD}}   &
\parbox{1.5cm}{\textbf{Ages \\(Years)}}\\
\midrule
\addlinespace
\parbox{4.8cm}{California Institute of Technology$^*$} & 
\parbox{1.cm}{38} &
\parbox{1.7cm}{1.0$\times$1.0$\times$1.0}   &
\parbox{1.3cm}{19/19}   &
\parbox{1.5cm}{17.0-56.2}\\
\addlinespace
\parbox{4.8cm}{Carnegie Mellon University$^*$} & 
\parbox{1.cm}{27} &
\parbox{1.7cm}{1.0$\times$1.0$\times$1.0}   &
\parbox{1.3cm}{13/14}    &
\parbox{1.5cm}{19-40}\\

\addlinespace
\parbox{4.8cm}{Kennedy Krieger Institute$^*$} & 
\parbox{1.cm}{55} &
\parbox{1.7cm}{1.0$\times$1.0$\times$1.0}   &
\parbox{1.3cm}{33/22}    &
\parbox{1.5cm}{8.0-12.8}\\

\addlinespace
\parbox{4.8cm}{Ludwig Maximilians 
University Munich$^*$} & 
\parbox{1.cm}{57} &
\parbox{1.7cm}{1.0$\times$1.0$\times$1.0}   &
\parbox{1.3cm}{33/24}    &
\parbox{1.5cm}{7-58}\\

\addlinespace
\parbox{4.8cm}{NYU Langone Medical Center$^*$} & 
\parbox{1.cm}{184} &
\parbox{1.7cm}{1.3$\times$1.0$\times$1.3}   &
\parbox{1.3cm}{105/79}    &
\parbox{1.5cm}{6.5-39.1}\\

\addlinespace
\parbox{4.8cm}{Olin, Institute of Living
at Hartford Hospital$^*$} & 
\parbox{1.cm}{36} &
\parbox{1.7cm}{1.0$\times$1.0$\times$1.0}   &
\parbox{1.3cm}{16/20}    &
\parbox{1.5cm}{10-24}\\

\addlinespace
\parbox{4.8cm}{Oregon Health and
Science University} & 
\parbox{1.cm}{28} &
\parbox{1.7cm}{1.0$\times$1.0$\times$1.1}   &
\parbox{1.3cm}{15/13}    &
\parbox{1.5cm}{8.0-15.2}\\

\addlinespace
\parbox{4.8cm}{San Diego State University$^*$} & 
\parbox{1.cm}{36} &
\parbox{1.7cm}{1.0$\times$1.0$\times$1.1}   &
\parbox{1.3cm}{22/14}    &
\parbox{1.5cm}{8.7-17.2 }\\

\addlinespace
\parbox{4.8cm}{Social Brain Lab
BCN NIC UMC Groningen$^*$} & 
\parbox{1.cm}{30} &
\parbox{1.7cm}{1.0$\times$1.0$\times$1.1}   &
\parbox{1.3cm}{15/15}    &
\parbox{1.5cm}{20-64 }\\

\addlinespace
\parbox{4.8cm}{Stanford University$^*$} & 
\parbox{1.cm}{40} &
\parbox{1.7cm}{0.86$\times$1.5$\times$0.86}   &
\parbox{1.3cm}{20/20}    &
\parbox{1.5cm}{7.5-12.9 }\\

\addlinespace
\parbox{4.8cm}{Trinity Centre for Health Sciences} & 
\parbox{1.cm}{49} &
\parbox{1.7cm}{1.0$\times$1.0$\times$1.0}   &
\parbox{1.3cm}{25/24}    &
\parbox{1.5cm}{12.0-25.9 }\\

\addlinespace
\parbox{4.8cm}{University of California,
Los Angeles: Sample 1$^*$} & 
\parbox{1.cm}{82} &
\parbox{1.7cm}{1.0$\times$1.0$\times$1.2}   &
\parbox{1.3cm}{33/49}    &
\parbox{1.5cm}{8.4-17.9}\\

\addlinespace
\parbox{4.8cm}{University of California,
Los Angeles: Sample 2} & 
\parbox{1.cm}{27} &
\parbox{1.7cm}{1.0$\times$1.0$\times$1.2}   &
\parbox{1.3cm}{14/13}    &
\parbox{1.5cm}{9.8-16.5 }\\

\addlinespace
\parbox{4.8cm}{University of Leuven: Sample 1$^*$} & 
\parbox{1.cm}{29} &
\parbox{1.7cm}{0.98$\times$0.98$\times$1.2}   &
\parbox{1.3cm}{15/14}    &
\parbox{1.5cm}{18-32}\\

\addlinespace
\parbox{4.8cm}{University of Leuven: Sample 2$^*$} & 
\parbox{1.cm}{35} &
\parbox{1.7cm}{0.98$\times$0.98$\times$1.2}   &
\parbox{1.3cm}{20/15}    &
\parbox{1.5cm}{12.1-16.9}\\

\addlinespace
\parbox{4.8cm}{University of Michigan: Sample 1} & 
\parbox{1.cm}{110} &
\parbox{1.7cm}{-$\times$-$\times$1.2}   &
\parbox{1.3cm}{55/55}    &
\parbox{1.5cm}{8.2-19.2}\\

\addlinespace
\parbox{4.8cm}{University of Michigan: Sample 2$^*$} & 
\parbox{1.cm}{35} &
\parbox{1.7cm}{-$\times$-$\times$1.2}   &
\parbox{1.3cm}{22/13}    &
\parbox{1.5cm}{12.8-28.8 }\\

\addlinespace
\parbox{4.8cm}{University of Pittsburgh
School of Medicine$^*$} & 
\parbox{1.cm}{57} &
\parbox{1.7cm}{1.1$\times$1.1$\times$1.1}   &
\parbox{1.3cm}{27/30}    &
\parbox{1.5cm}{9.3-35.2 }\\

\addlinespace
\parbox{4.8cm}{University of Utah
School of Medicine$^*$} & 
\parbox{1.cm}{101} &
\parbox{1.7cm}{1.0$\times$1.0$\times$1.2}   &
\parbox{1.3cm}{43/58}    &
\parbox{1.5cm}{8.8-50.2}\\

\addlinespace
\parbox{4.8cm}{Yale Child Study Center} & 
\parbox{1.cm}{56} &
\parbox{1.7cm}{1.0$\times$1.0$\times$1.0}   &
\parbox{1.3cm}{28/28}    &
\parbox{1.5cm}{7.0-17.8}\\

\addlinespace
\bottomrule
\end{tabular}
\caption{Scan parameters and characteristics of sites included in the ABIDE dataset. An asterisk beside the site name indicates that data from this site were used for training.}
\label{tab:sites}
\end{table}

Unlike IBSR, the ABIDE dataset does not contain ground-truth segmentations of subcortical structures. Instead, we have used automatic segmentations obtained using the \emph{recon-all} pipeline\footnote{\url{http://surfer.nmr.mgh.harvard.edu/fswiki/recon-all}} of the \FreeSurfer{} 5.1 tool \cite{fischl2012freesurfer}, which are freely available at \url{http://fcon_1000.projects.nitrc.org/indi/abide/}. This pipeline involves the following steps: motion correction, intensity normalization, affine registration of volumes to the MNI305 atlas, skull-stripping, non-linear registration using the Gaussian Classifier Atlas (GCA), and brain parcellation. The outputs of this pipeline used in our study are the skull stripped, intensity normalized brain volumes in the unregistered subject space (i.e., \emph{brain.mgz} files) and the subcortical labelling of these volumes (i.e., \emph{aseg.mgz} files). Note that FreeSurfer’s registration and segmentation steps were shown to be robust to age-associated bias \cite{ghosh2010evaluating}.

For this dataset, the objectives of our experiment was to measure the impact of different imaging, demographic and clinical factors on the reliability of the proposed method. Another goal was to verify that our method could obtain segmentations similar to those of atlas-based approaches (e.g., the segmentation approach of \FreeSurfer{}), but in a fraction of the time. To measure the impact of age (and thus brain size) on our method's performance, we followed the methodology of \cite{aylward2002effects} and divided subjects into three non-overlapping groups: $<$13 years, 13 to 18 years, and $>$18 years. Furthermore, to account for potential structural differences related to autism, we further split each age group into two sub-groups, containing control and ASD subjects respectively. Lastly, to evaluate the robustness of our method in unseen cohorts, the resulting subject groups were again split based on whether the subject is from a site used in training or not. Note that, in the case of subjects from sites used in training, only subjects from the test set are considered (i.e., no training example is used while measure the segmentation performance). A summary of group configuration and training/testing distribution is presented in Table \ref{table:groupsConf}. To facilitate the presentation of results, each group is identified by a unique ID, from A to L. For instance, group A corresponds to control subjects less than 13 years of age, from sites used in training. Among this group, the data of 42 subjects were used for training our FCNN, and the data of 93 subjects for measuring its performance.

\begin{table}[htb!]
\setlength{\tabcolsep}{1pt}
\scriptsize
\centering
\renewcommand{\arraystretch}{1.25}
\begin{tabular}{|C{17mm}|C{8mm}|C{8mm}|C{8mm}|C{8mm}|C{8mm}|C{8mm}|C{7.5mm}|C{7.5mm}|C{7.5mm}|C{7.5mm}|C{7.5mm}|C{7.5mm}|}
\hline
 & \multicolumn{6}{c|}{\textbf{Site used in training}} & \multicolumn{6}{c|}{\textbf{Site NOT used in training}} \\ 
 \hline
\textbf{DX group}  & \multicolumn{3}{c|}{Control} & \multicolumn{3}{c|}{ASD} & \multicolumn{3}{c|}{Control} & \multicolumn{3}{c|}{ASD} \\ 
\hline
\textbf{Age group} & $<\!13$ & 13-18 & $>\!18$ & $<\!13$ & 13-18 & $>\!18$ & $<\!13$ & 13-18 & $>\!18$ & $<\!13$  & 13-18      & $>\!18$      \\ 
\hline 
\textbf{Subjects \newline train/test} & 42/93 & 46/87 & 62/98  & 0/133 & 0/113 & 0/144 & 0/65 & 0/53 & 0/16 & 0/62 & 0/48 & 0/11      \\
\hhline{|=|=|=|=|=|=|=|=|=|=|=|=|=|}
\textbf{Group ID} & A & B & C & D & E & F & G & H & I & J & K & L \\ \hline 
\end{tabular}
\caption{Configuration of subject groups used in the proposed experiments. The number of training and testing subjects included in each group is detailed in last row.}
\label{table:groupsConf}
\end{table}

We evaluated the segmentation of target subcortical brain structures by training and testing with data from different sites or age/diagnosis groups. For training, we considered 10 control subjects from 15 sites (indicated by an asterisk in Table \ref{tab:sites}), giving a total of 150 training examples. The reason for employing only control subjects is that ASD subjects may present structural abnormalities that are not representative of the population. Including these subjects in the training set might thus affect the generability of our segmentation model. For validation, we used a single subject per site, leading to a validation set composed of 15 examples. Segmentation performance was evaluated on remaining subjects from all sites.

\subsubsection{Implementation details} 
\label{sssec:architecture}

Selecting the network's architecture is a complex and problem-specific task, which can greatly affect the performance and computational efficiency of the solution. In this study, we investigate three different FCNN architectures. The first architecture, called $\CNNbase$, is composed of 3 convolutional layers with 25, 50 and 75 feature maps (i.e., filters), respectively, and a kernel size of $7\!\times\!7\!\times 7$. Three fully-connected layers are added after the last convolutional layer to model the relationship between features and class labels. The $\CNNbase$ architecture, depicted in Figure \ref{fig:CNNBase}, is employed as a baseline to generate ``standard'' or ``control'' segmentations. In the second architecture, denoted as $\CNNsingle$, each convolutional layer is replaced by three successive convolutional with smaller kernels of size $3\!\times\!3\!\times 3$. As mentioned in Section \ref{ssec:smallKernels}, this strategy allows having a deeper network with the same number of parameters. Finally, the third architecture, called $\CNNmulti$, corresponds to the multiscale FCNN of Figure \ref{fig:CNN_archit}, which was presented in Section \ref{ssec:combiningFeatures}.

All three architectures have three fully-connected layers, composed of 400, 200 and 150 hidden units respectively. These layers are followed by a final classification layer, which outputs the probability maps for each of the 9 classes: 8 for each of the subcortical structures (left and right) and one for the background. The $\CNNmulti$ architecture proposed in this paper is thus composed of 13 layers in total, with the following layout: 9 convolutional layers, 3 fully-connected layers, and the classification layer. Furthermore, the number of kernels in each convolutional layer (from first to last) is as follows: 25, 25, 25, 50, 50, 50, 75, 75 and 75.

The optimization of network parameters is performed with stochastic gradient descent (SGD), using cross-entropy as cost function. However, since our network employs 3D convolutions, and due to the large sizes of MRI volumes, dense training cannot be applied to whole volumes. Instead, volumes are split into $B$ smaller segments, which allows dense inference in our hardware setting. Let $\theta$ be the network parameters (i.e., convolution weights and biases), and denote as $\mathcal{L}$ the set of ground-truth labels such that $L^v_s \in \mathcal{L}$ is the label of voxel $v$ in the $s$-th image segment. Following \cite{kamnitsas2016efficient}, we defined the cost function as
\begin{equation}
\qquad \qquad  J(\theta; \mathcal{L}) \ = \ 
    -\frac{1}{B\!\cdot\!V} \sum^{B}_{s=1} \sum^{V}_{v=1} \log \, p_{L^v_s}(X_v),
\end{equation}
where $p_c(X_v)$ is the output of the classification layer for voxel $v$ and class $c$. In \cite{kamnitsas2016efficient}, Kamnitsas et al. found that increasing the size of input segments in training leads to a higher performance, but this performance increase stops beyond segment sizes of \vold{25}. In their network, using this segment size for training, score maps at the classification stage were of size \vold{9}. Since our architecture is one layer deeper, and to keep the same score map sizes, we set the segment size in our network to \vold{27}.

Deep CNNs are usually initialized by assigning random normal-distributed values to kernel and bias weights. As demonstrated in \cite{simonyan2014very}, initializing weights with fixed standard deviations may lead to poor convergence. To overcome this limitation, we adopted the strategy proposed in \cite{he2015delving}, and used in \cite{kamnitsas2016efficient} for segmentation, that allows very deep architectures (e.g., 30 convolutional or fully-connected layers) to converge rapidly. In this strategy, weights in layer $l$ are initialized based on a zero-mean Gaussian distribution of standard deviation  $\sqrt{2/n_l}$, where $n_l$ denotes the number of connections to units in that layer. For example, in the first convolutional layer of Figure \ref{fig:CNN_archit}, the input is composed of single-channel (i.e., grey level) image segments and kernels have a size of \vold{3}, therefore the standard deviation is equal to $\sqrt{2/(1\times3\times3\times3)} = 0.2722$.

Our 3D FCNNs were initially trained for 50 epochs, each one composed of 20 subepochs. At each subepoch, a total of 500 samples were randomly selected from the training image segments, and processed in batches of size 5. However, we observed that the performance of the trained network on the validation set did not improve after 30 epochs, allowing us to terminate the training process at this point. As other important meta-parameters, the training momentum was set to 0.6 and the initial learning rate to 0.001, being reduced by a factor of 2 after every 3 epochs. Note that instead of an adaptive strategy for learning rate, we used step decay and monitored the evolution of the cost error during training. We observed that it followed a decreasing curve, with and exponential shape and small increasing/decreasing slopes. Therefore, we considered that decreasing the learning rate at certain epochs was sufficient to properly train the network.

To implement our network, we adapted the 3D FCNN architecture of Kamnitsas et al. \cite{kamnitsas2016efficient}. Their architecture was developed using Theano, a CPU and GPU mathematical compiler for implementing deep learning models \cite{bergstra2010theano}. The PC used for training is an Intel(R) Core(TM) i7-6700K 4.0GHz CPU, equipped with a NVIDIA GeForce GTX 960 GPU with 2 GB of memory. Training our network took a little over 2 hours per epoch, and around 2 days and a half for the fully trained CNN. The source code of our implementation is publicly available on \sloppy\burl{https://github.com/josedolz/LiviaNET}.

\subsubsection{Evaluation}
\label{sssec:evaluation}

Various comparison metrics exist to evaluate the accuracy of segmentation methods. Although volume-based metrics, such as Dice similarity coefficient (DSC) \cite{dice1945measures}, have been broadly used to compare segmentation results, they are fairly insensitive to the precise contour of segmented regions, which only has small impact on the overall volume. However, two segmentations with a high spatial overlap may exhibit clinically relevant differences in their boundaries. To measure such differences, distance-based metrics such as the Modified Hausdorff distance (MHD) are typically used.

\paragraph{\textbf{Dice similarity coefficient}}

Let $V_\mr{ref}$ and $V_\mr{auto}$ denote the binary reference segmentation and the automatic segmentation, respectively, of a given tissue class for a given subject. The DSC is then defined as
\begin{equation}
	\qquad \qquad \qquad \qquad 
	    \mr{DSC}\big(V_\mr{ref}, V_\mr{auto} \big) \ = \ \frac{2 \mid V_\mr{ref} \cap V_\mr{auto}\mid} {\mid V_\mr{ref}\mid +\mid V_\mr{auto}\mid}
\end{equation}
DSC values are comprised in the $[0,1]$ range, where 1 indicates perfect overlapping and 0 represents no overlapping at all.

\paragraph{\textbf{Modified Hausdorff distance}}

Let $P_\mr{ref}$ and $P_\mr{auto}$ denote the sets of voxels within the reference segmentation and the automatic one, respectively. The MHD can be then defined as
\begin{equation}
	\qquad  \mr{MHD}\big(P_\mr{ref}, P_\mr{auto} \big) \ = \ 
	    \max \Big\{ d(P_\mr{ref},P_\mr{auto}), d(P_\mr{auto},P_\mr{ref}) \Big\},
\end{equation}
where $d(P,P')$ is the maximum distance between a voxel in $P$ and its nearest voxel in $P'$. In this case, smaller values indicate higher proximity between two point sets, and thus a better segmentation.

\section{Results}
\label{ssec:results}

We first test our segmentation method on the IBSR dataset, which has been used in numerous studies on subcortical parcellation. In Section \ref{ssec:ABIDEDatRes}, we then measure the benefits of having a deeper network with smaller kernels and using multiscale features, as well as evaluate the impact of various acquisition, demographics, and clinical factors, by applying our $\CNNbase$, $\CNNsingle$ and $\CNNmulti$ architectures on the ABIDE dataset. Finally, in Section \ref{ssec:IBSR_ABIDE}, we demonstrate the cross-dataset generalization  of our method by evaluating on the IBSR dataset the FCNN trained using the ABIDE dataset.

For notation simplicity, we now on denote brain structures by their first two characters, indicating within parenthesis their location, i.e left (L) or right (R) hemisphere. For example, the caudate in the left brain side will be referred to as Ca(L).

\subsection{Evaluation on the IBSR dataset}
\label{ssec:IBSRDatRes}

\begin{figure}[h!]
\begin{center} 
     \mbox{
        \includegraphics[height=0.35\textwidth]{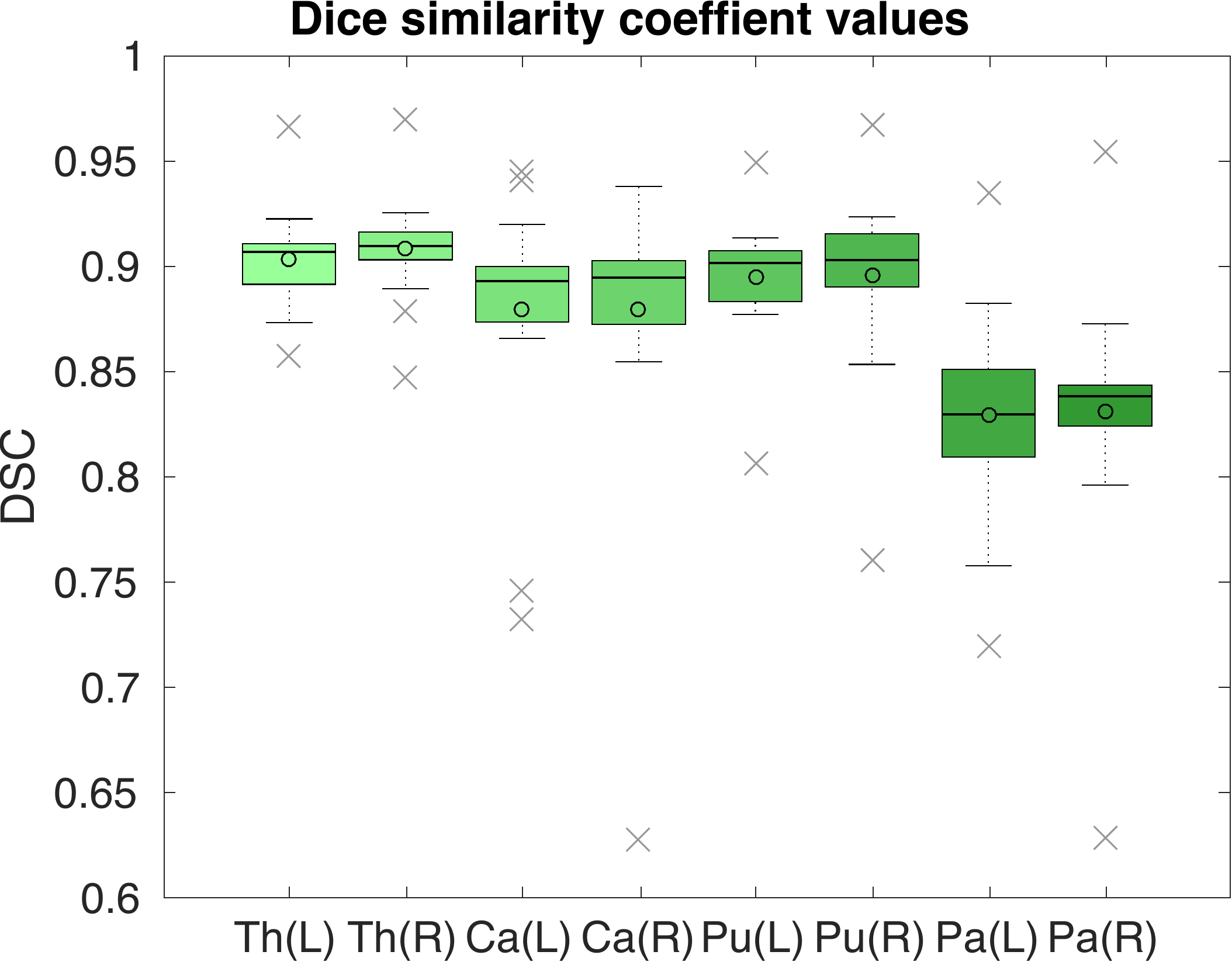}
        
        \hspace{2mm}
        
        \includegraphics[height=0.35\textwidth]{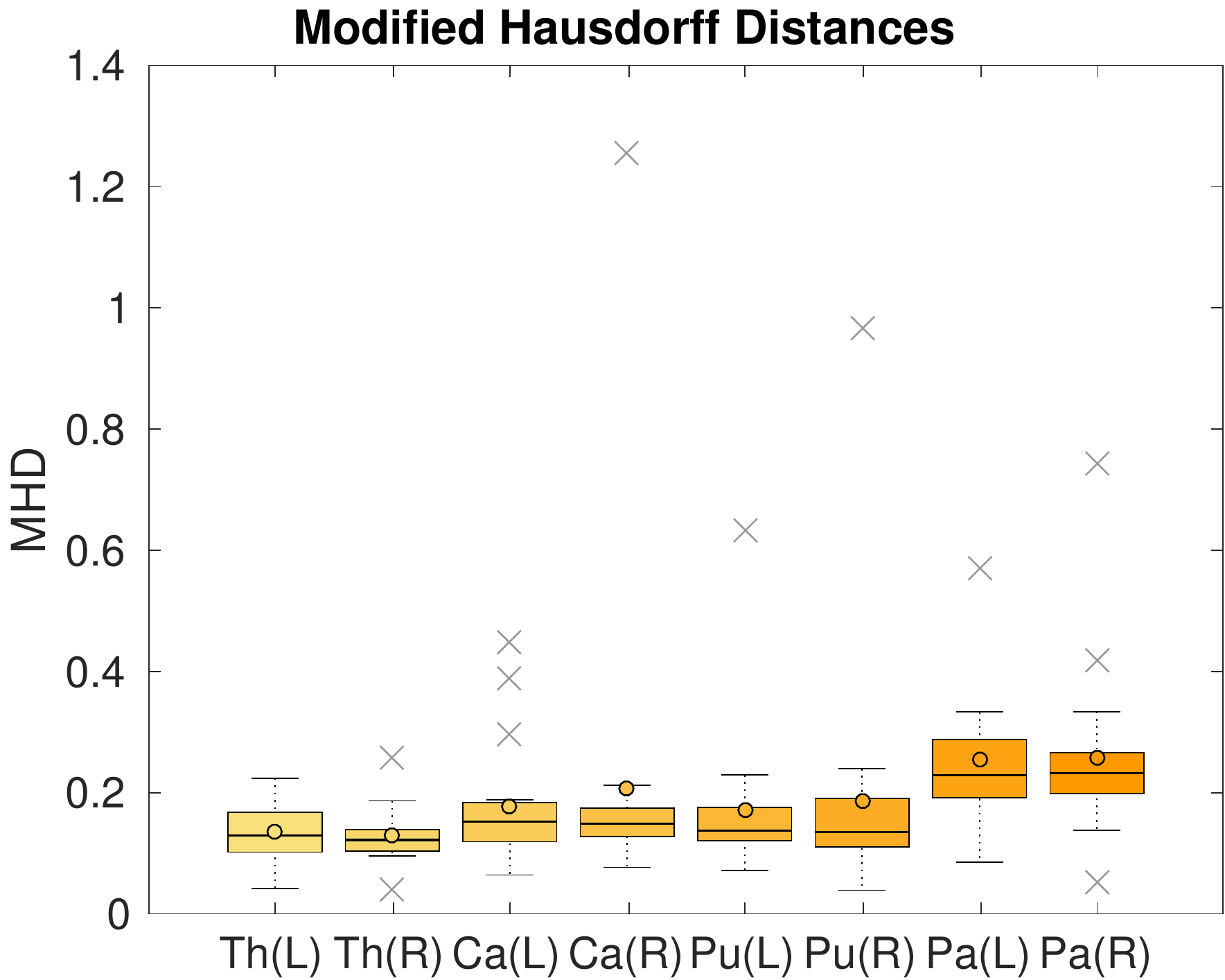}
        }
        \caption{Segmentation accuracy (mean DSC and MHD) obtained, for each brain structure, by the $\CNNmulti$ architecture on subjects of the IBSR dataset.}
        \label{fig:res_IBSR}
\end{center}        
\end{figure}

Figure \ref{fig:res_IBSR} shows the segmentation accuracy of the proposed $\CNNmulti$ architecture, in terms of the DSC and MHD, obtained for the target left- and right-side brain structures. We see that the segmentation of the pallidum, both left and right, was significantly less accurate than other structures (i.e., thalamus, caudate and putamen), likely due to the smaller size of this brain structure. Furthermore, we observe that the segmentation of all four subcortical structures is slightly more accurate in the right hemisphere, although the differences are not statistically significant following a Wilcoxon signed-rank test.

In light of the various studies using the IBSR dataset as segmentation benchmark, the results obtained by our method are state-of-the-art (Table \ref{tab:sumState}). Specifically, when comparing against the recent work of Shakeri et al. \cite{shakeri2016sub}, which used a 2D FCNN and the same validation methodology, our method achieved DSC improvements ranging from 5\% (in the thalamus) to 13\% (in the caudate).

To demonstrate that the proposed approach actually learns from training data, we also measured the performance of a simple majority voting technique, using the same leave-one-out-cross-validation strategy. In this technique, each voxel of the volume to segment is mapped to the most frequent class in corresponding voxels of training volumes. Applying this technique to all IBSR subjects gave mean DSC values of 0.83, 0.69, 0.74 and 0.75 for the thalamus, caudate, putamen and pallidum, respectively.

\subsection{Evaluation on the ABIDE dataset}
\label{ssec:ABIDEDatRes}

Table \ref{table:ABIDEThreeConf} gives the mean DSC and mean MHD obtained by the $\CNNbase$, $\CNNsingle$ and $\CNNmulti$ architectures on all test examples of the the ABIDE dataset. Recall that these accuracy measures were computed using the labels obtained from \FreeSurfer{}, since ground-truth segmentations were not available. 
We first observe that having a deeper network, via smaller kernels, increases the segmentation performance in both metrics. In a one-sided A one-sided non-parametric statistical test can handle non-Gaussian data distributions, and accommodates the following alternative hypothesis: The architecture is \emph{better} than the baseline. Wilcoxon signed-rank test or $t$-test, the mean DSC and MHD of $\CNNsingle$ is statistically better (i.e., higher for DSC and lower for MHD) than $\CNNbase$, with $p < 0.01$. Likewise, when features extracted at intermediate layers are fed into the first fully-connected layer, the proposed $\CNNmulti$ network generated more reliable segmentations, both in terms DSC and MHD. These results are also statistically significant, with $p < 0.01$, in a Wilcoxon signed-rank test or $t$-test.

\begin{table}[h!]
\begin{center}
\begin{footnotesize}
\renewcommand{\arraystretch}{1.2}
\begin{tabular}{llccc}
\toprule
    & \textbf{Structure} & $\bm{\CNNbase}$ & $\bm{\CNNsingle}$ & $\bm{\CNNmulti}$  \\ 
    \midrule
\multirow{8}{*}{Mean DSC} & \multirow{2}{*}{Thalamus} & 0.8987 & 0.9039 & \textbf{0.9156} \\
&  & (0.002) & (0.0052) & (0.0012) \\
\cmidrule{2-5}
& \multirow{2}{*}{Caudate} & 0.8979 & 0.9011 & \textbf{0.9073} \\
&  & (0.0012) & (0.0002) & (0.0021)\\
\cmidrule{2-5}
& \multirow{2}{*}{Putamen} & 0.8909 & 0.8992 & \textbf{0.9041} \\
&  & (0.0017) & (0.0009) & (0.0014) \\
\cmidrule{2-5}
& \multirow{2}{*}{Pallidum} & 0.8381 & 0.8497 & \textbf{0.8621} \\
&  & (0.0096) & (0.0096) & (0.0095) \\
\midrule
\multirow{8}{*}{Mean MHD} & \multirow{2}{*}{Thalamus} & 0.1487 & 0.1462 & \textbf{0.1405} \\
&  & (0.0087) & (0.0153) & (0.0002) \\
\cmidrule{2-5}
& \multirow{2}{*}{Caudate} & 0.1710 & 0.1583 & \textbf{0.1557} \\
&  & (0.0154) & (0.0213) & (0.0165) \\
\cmidrule{2-5}
& \multirow{2}{*}{Putamen} & 0.2074 & 0.1742 & \textbf{0.1706} \\
&  & (0.0296) & (0.0467) & (0.0296) \\
\cmidrule{2-5}
& \multirow{2}{*}{Pallidum} & 0.2487 & 0.2305 & \textbf{0.2232} \\
&  & (0.0205) & (0.0208) & (0.0232) \\
\bottomrule
\end{tabular}
\end{footnotesize}
\caption{Mean DSC and MHD (standard deviation between brackets), obtained by the three tested FCNN architectures on the ABIDE dataset. Bold font numbers indicate the best result among the three architectures.}
\label{table:ABIDEThreeConf}
\end{center}
\end{table}

\begin{figure}[h!]
     \begin{center}
     \mbox{
        \includegraphics[width=0.485\textwidth]{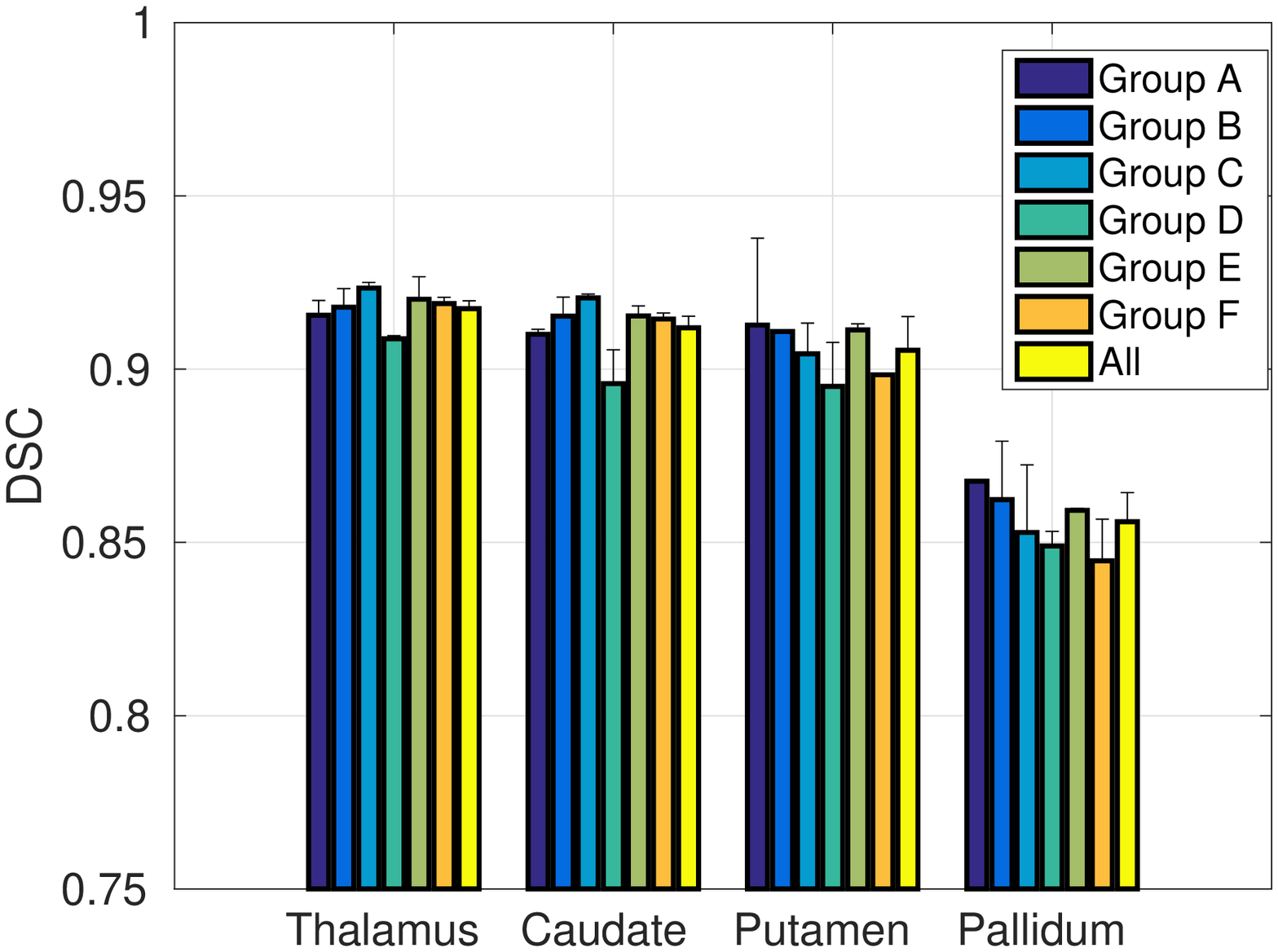}
        \includegraphics[width=0.485\textwidth]{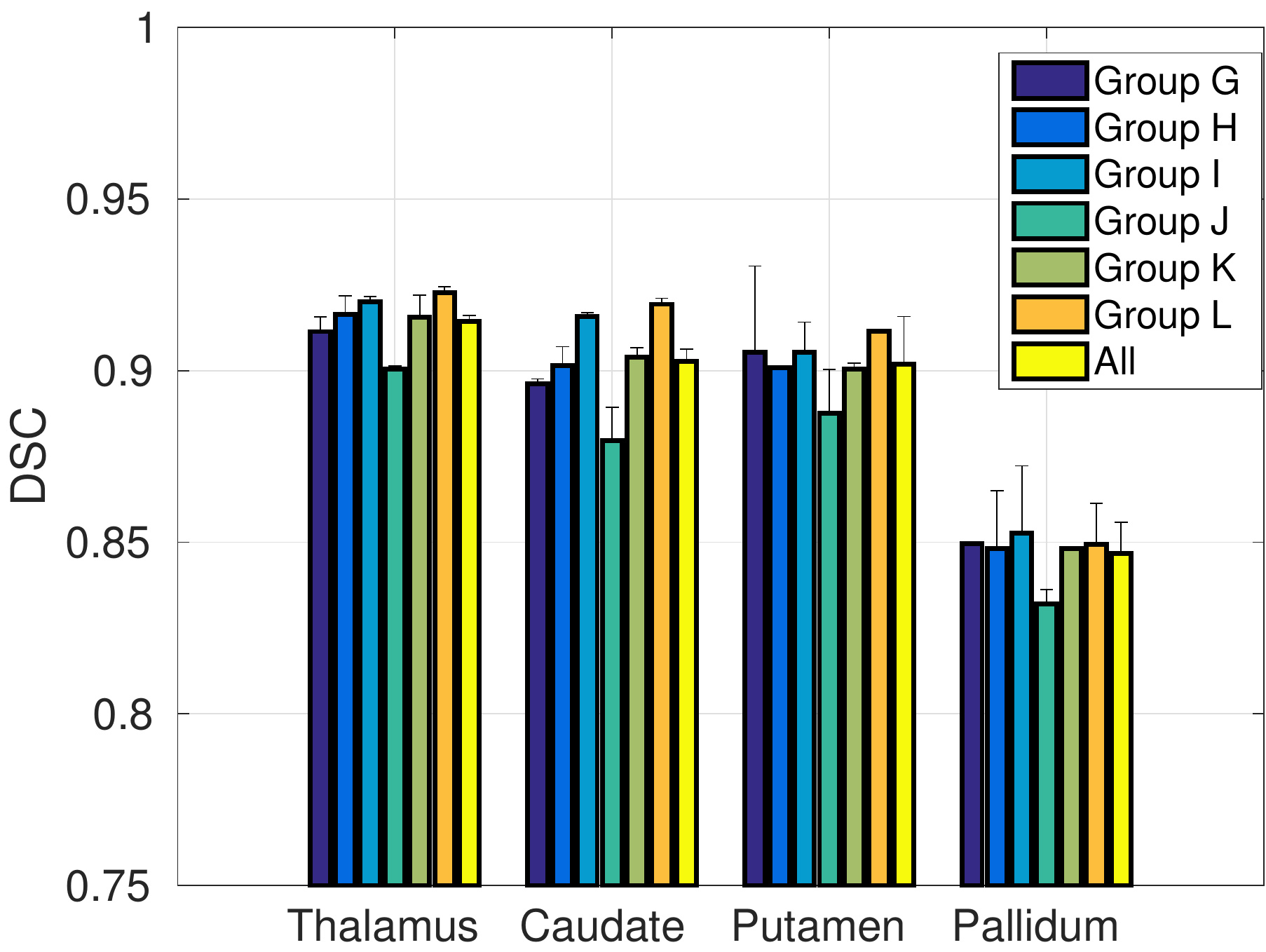}
     }
     
     \vspace{1mm}
     
     \mbox{
        \shortstack{\includegraphics[width=0.485\textwidth]{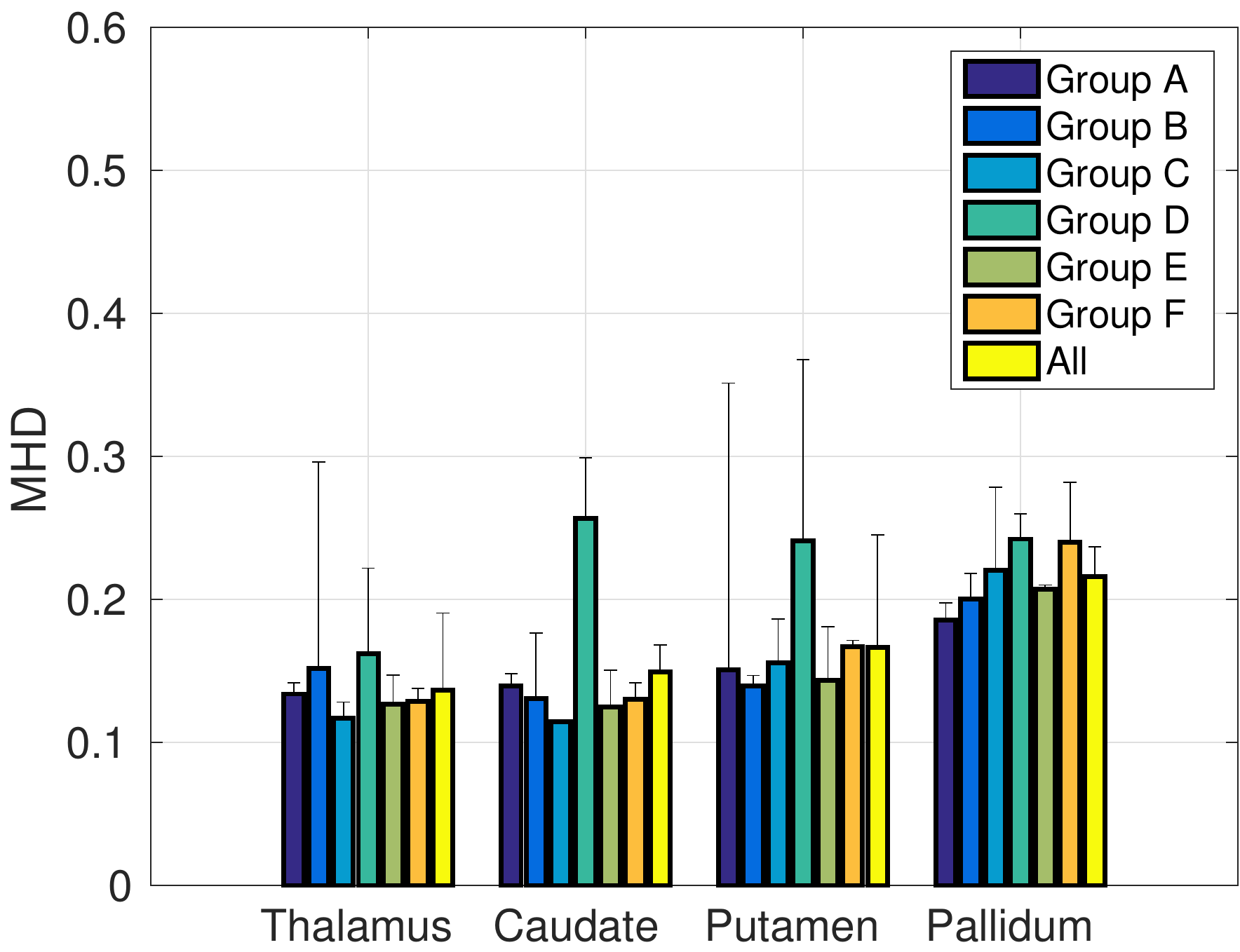} \\
            a) Data employed in training}
        \shortstack{\includegraphics[width=0.485\textwidth]{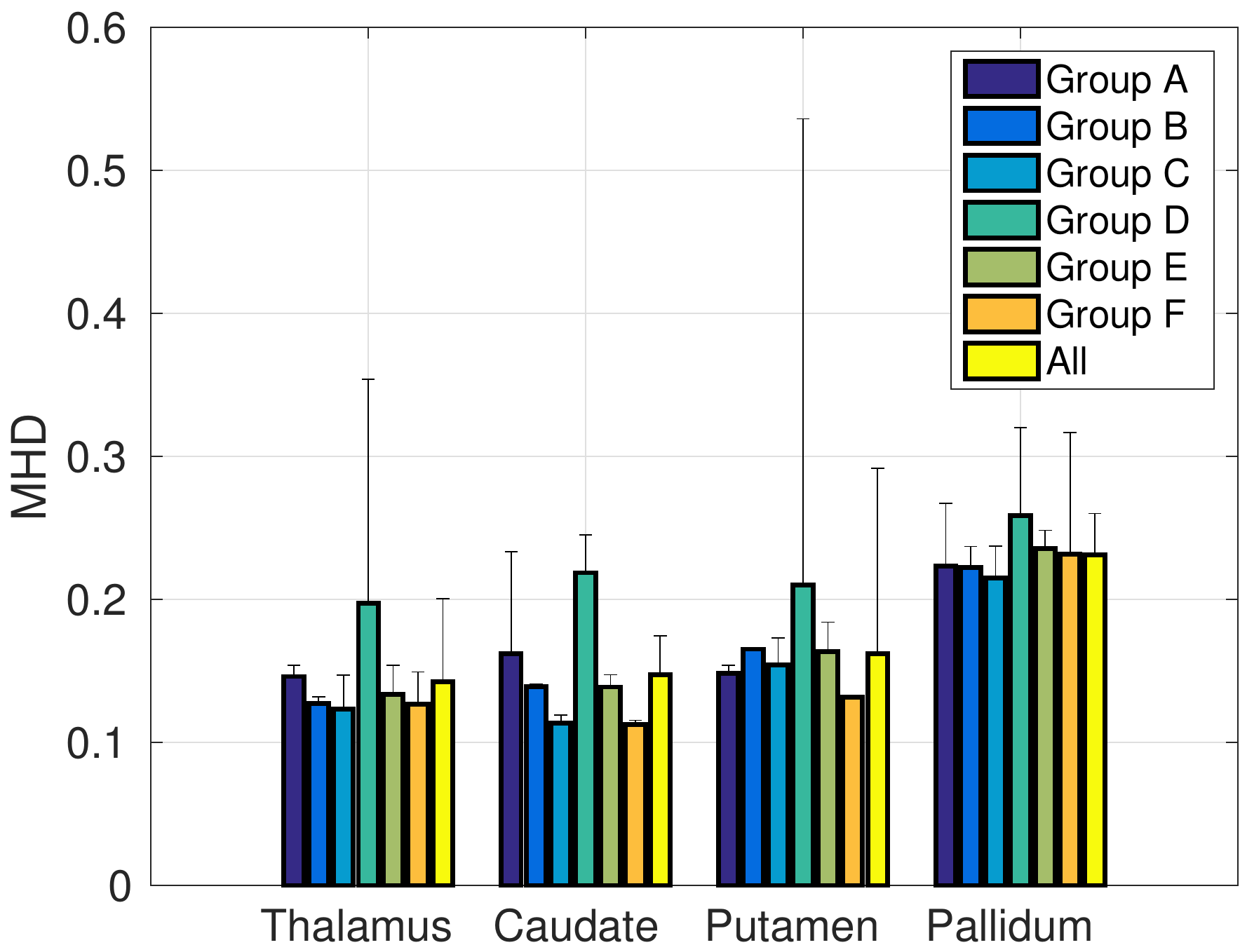} \\
            b) Unseen data}
        }
        \caption{Mean DSC and MHD values obtained for subject data used during training, and for unseen data.}
        \label{fig:dsc1}
\end{center}        
\end{figure}

Figure \ref{fig:dsc1} plots the mean DSC and MHD values obtained by our $\CNNmulti$ architecture for each of the subject groups in Table \ref{table:groupsConf}. These values are grouped by subcortical structure of interest, i.e., thalamus, caudate, putamen and pallidum. For each structure, an additional bar is added, giving the mean DSC and MHD obtained on subjects of all groups together. Across all subject groups, the segmentations produced by our 3D FCNN achieved mean DSC values above 0.90 for all structures except the pallidum, which had a mean DSC of 0.85. Likewise, mean MHD values were below 0.25 mm in all subject groups and for all four subcortical structures. These results are consistent with those obtained for the IBSR dataset.

Analyzing the results obtained using data from sites considered in training (groups A-F), we observe that mean DSC values obtained for control subjects (groups A-C) are usually higher than for ASD subjects (groups D-E). For instance, putamen segmentation in control subjects less than $13$ years old yielded a mean DSC of 0.9127, compared to 0.9055 for ASD subjects in the same age group. The same trend is seen for distance similarities, for example in the caudate, where a mean MHD of 0.1397 was obtained for control subjects, versus 0.2568 for ASD subjects. These results illustrate that physiological differences related to autism, especially in young subjects, can have a small impact on segmentation accuracy.

\begin{figure}[h!]
     \begin{center}
     \mbox{
        \includegraphics[width=0.485\textwidth]{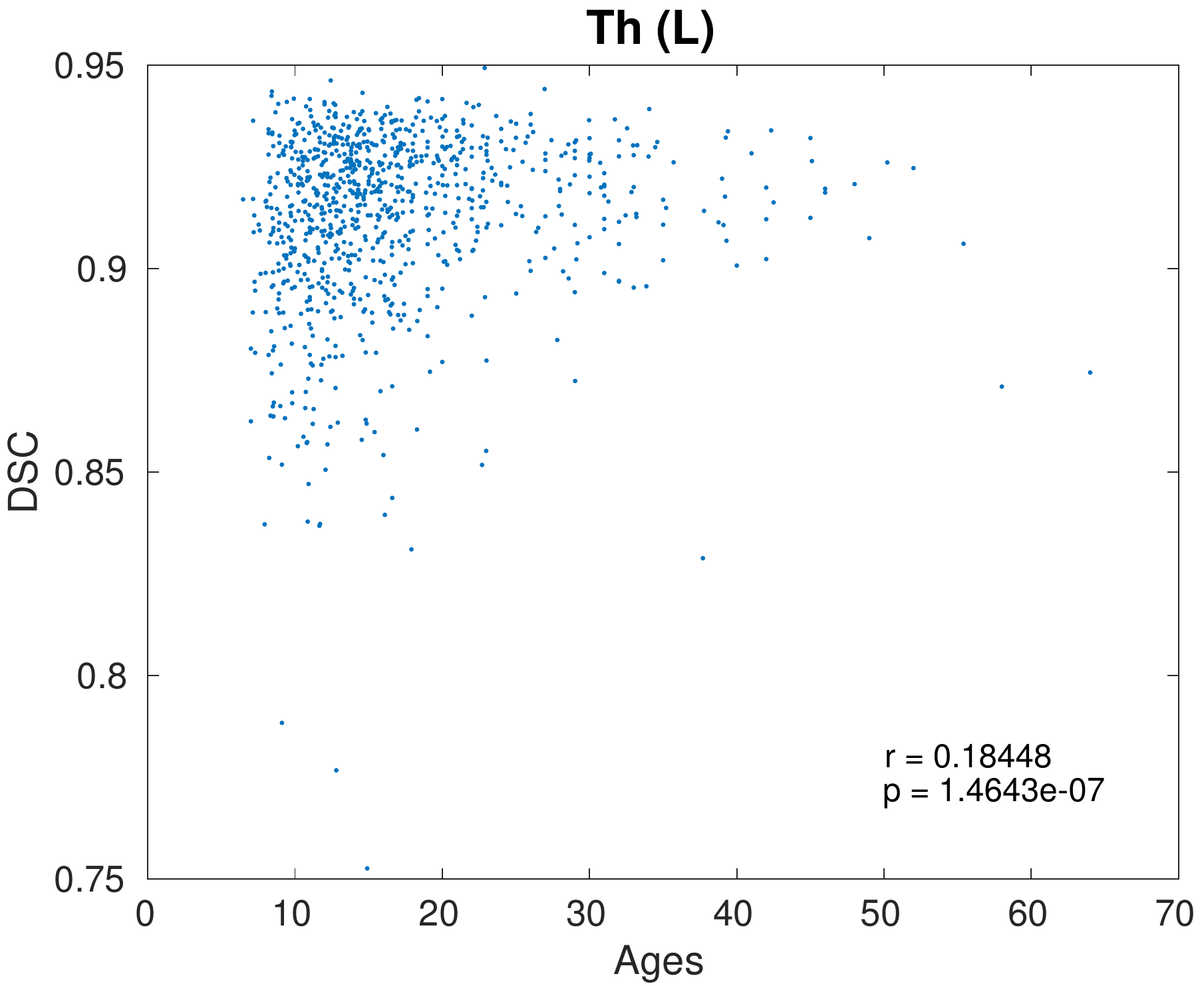}
        \includegraphics[width=0.485\textwidth]{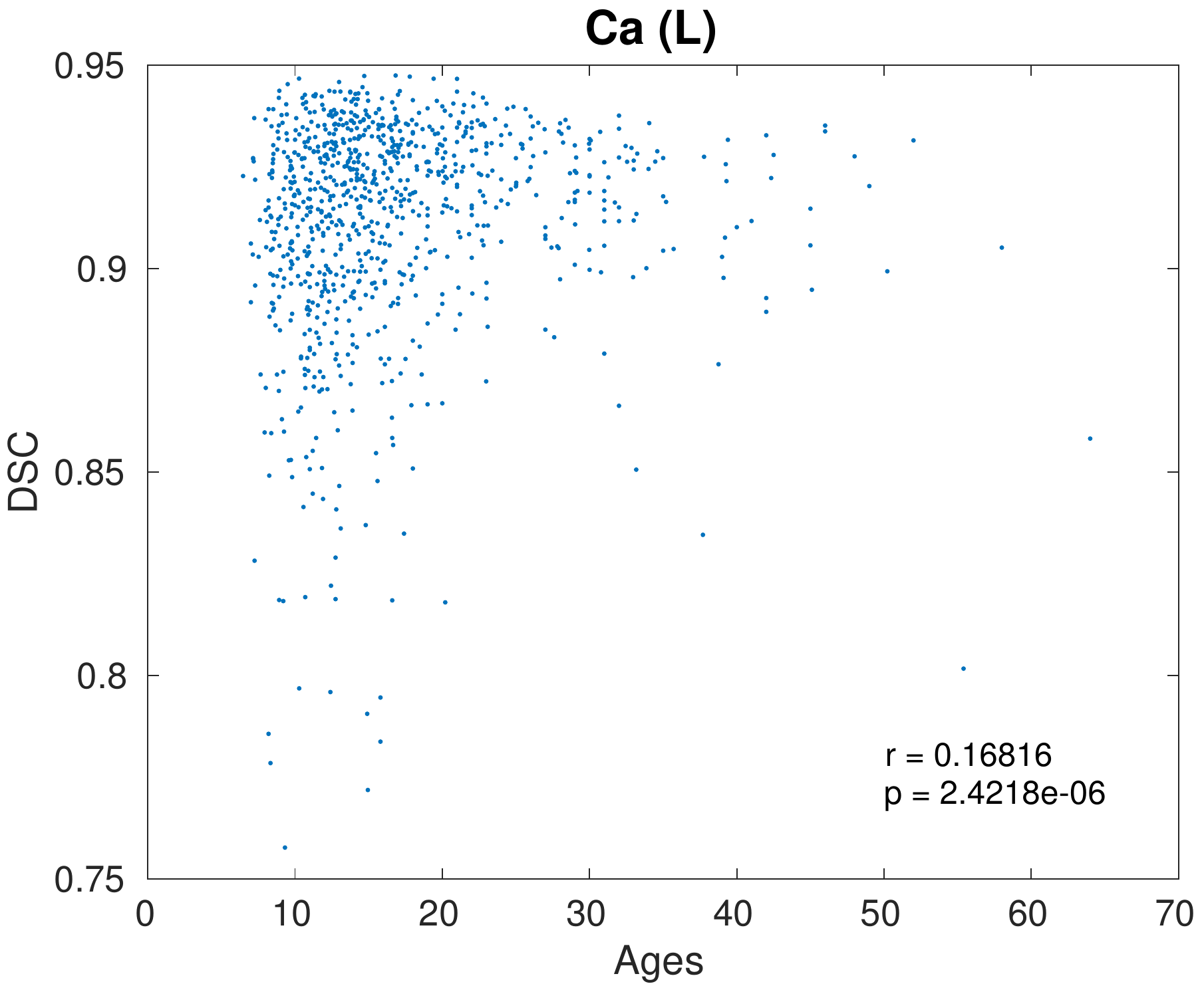}
     }
     \mbox{
        \includegraphics[width=0.485\textwidth]{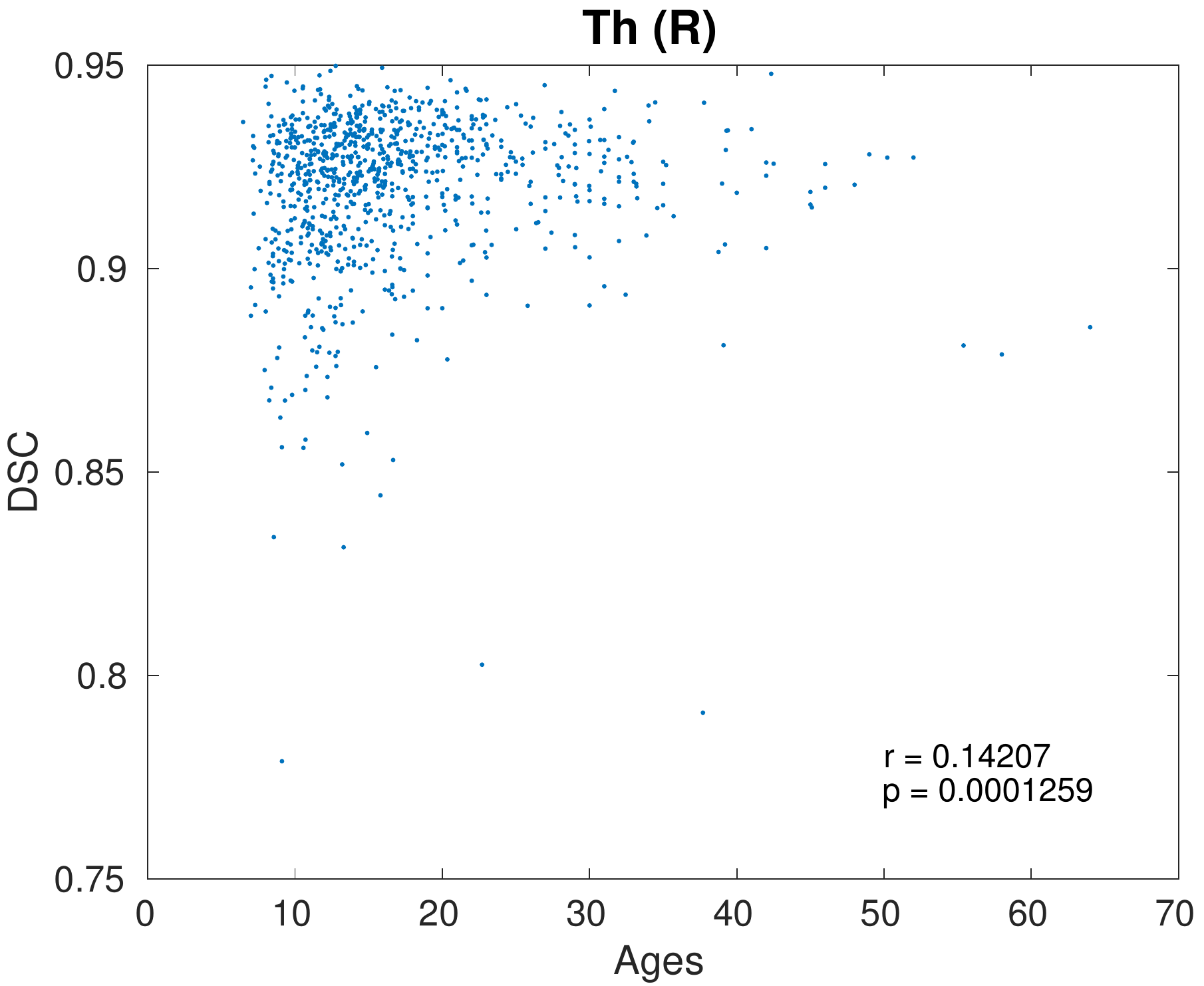}
        \includegraphics[width=0.485\textwidth]{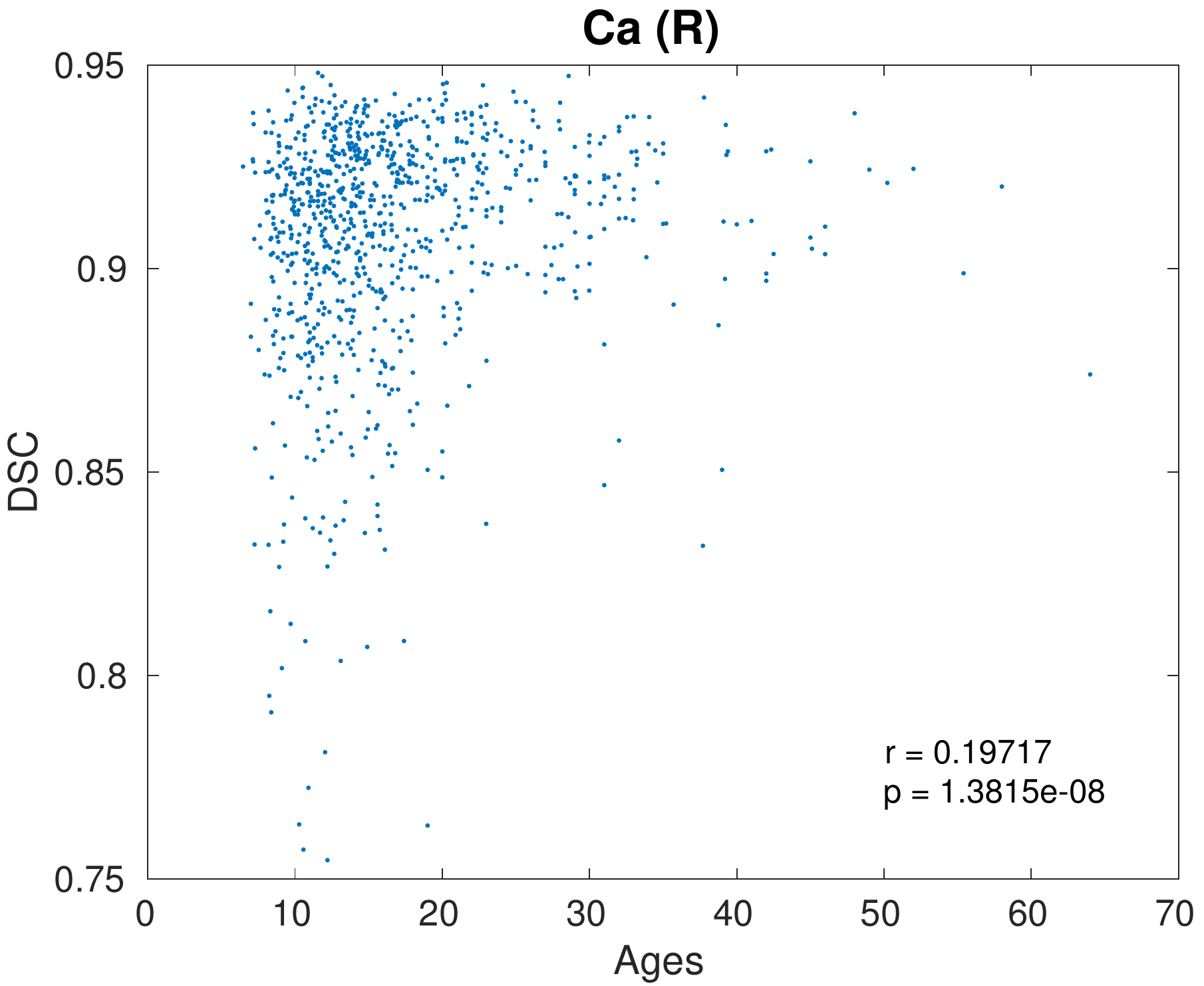}
     }
     
        \caption{Scatter plots of left and right thalamus and caudate segmentation performance regarding DSC and subject age, where the Spearman rank correlation coefficient $r$ and corresponding $p$-values are given for each plot. From these plots, a weak but statistically significant correlation between performance and subject age is observed.}
        \label{fig:scat}
\end{center}        
\end{figure}

Looking at the impact of subject age on results, it can be seen that the segmentation of the thalamus and caudate improves as the subject gets older, in both control and ASD subjects. The relationship between subject age and segmentation accuracy in these structures is further illustrated in Figure \ref{fig:scat}, which gives the scatter plot of DSC versus age in the left/right thalamus and caudate, considering all control and ASD subjects together. In each plot, the Spearman rank correlation coefficient and corresponding $p$-value are given as variables $r$ and $p$. Note that $p$-values have been corrected using the Bonferroni procedure, to account for the multiple comparisons (8 structures). We notice a weak but statistically significant correlation, with $p < 0.01$, validating our previous observation. It is also worth noting a greater variance in accuracy occurring for younger subjects, most of the low accuracy values observed for ages less than 20 years old. This is consistent with the fact that the brain is continuously developing until adulthood, and suggests that the physiological variability of younger subjects may not be completely captured while training the network.

The same patterns can be observed when segmenting subjects from sites not used in training (groups G-L). Particularly noticeable is the relationship between age and accuracy, which can be seen in all structures, and in both control and ASD subjects. Comparing with results obtained on data from sites used in training, we find no statistically significant difference in accuracy (DSC or MHD), for any brain structure. This suggests that the proposed method can generalize to acquisition protocols and imaging parameters not seen in training. Overall the results of these experiments illustrate that our method is robust to various clinical, demographics and site-related factors.    

\begin{figure}[ht!]
     \begin{center}
     \textbf{Site used in training}
        \vspace{1mm}
     
        \mbox{
            \shortstack{\FreeSurfer{} \\  
                \hspace{0.25pt} \includegraphics[width=0.35\textwidth]{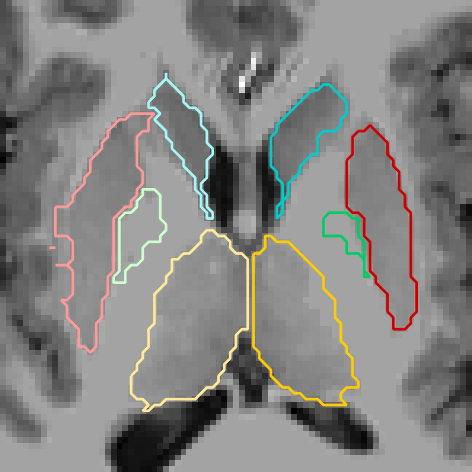}}
            \shortstack{Our CNN \\ 
                \includegraphics[width=0.35\textwidth]{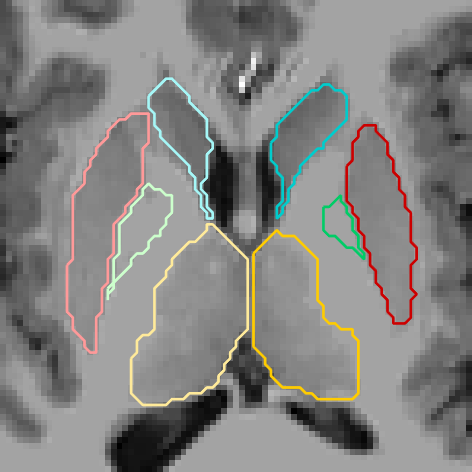}}
        }
        \mbox{
            \includegraphics[width=0.35\textwidth]{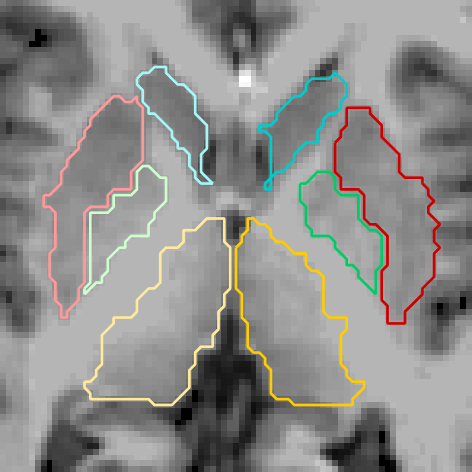}
            \includegraphics[width=0.35\textwidth]{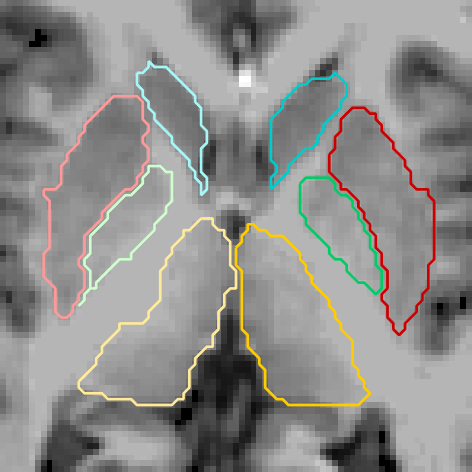}
        }
        
        \vspace{1mm}
        
        \mbox{
            \includegraphics[width=0.35\textwidth]{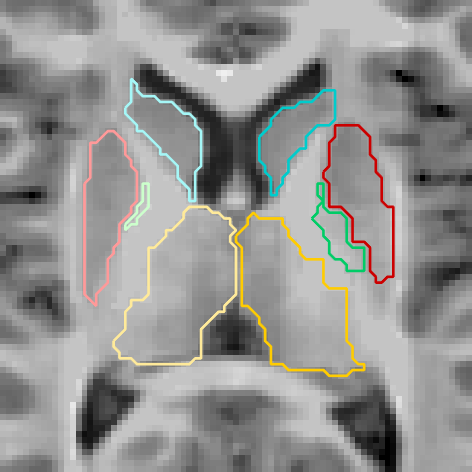}
            \includegraphics[width=0.35\textwidth]{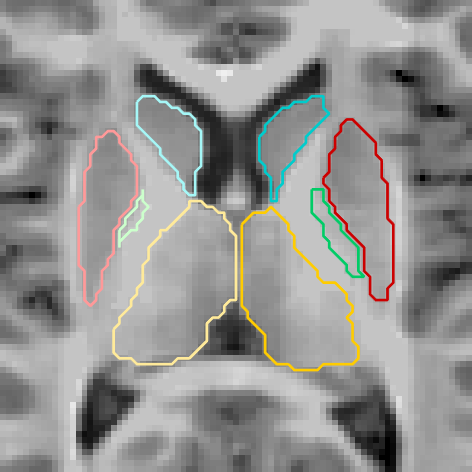}
        }
    \end{center}
    \caption{Visual examples of our 3D FCNN architecture compared with the standard references contoured by \FreeSurfer{}, for three test subjects from sites used in training.}%
   \label{fig:subfiguresFirstDataset}
\end{figure}

\begin{figure}[ht!]
     \begin{center}
          \textbf{Site NOT used in training}
          
          \vspace{1mm}
     
        \mbox{
            \shortstack{\FreeSurfer{} \\  
                \includegraphics[width=0.35\textwidth]{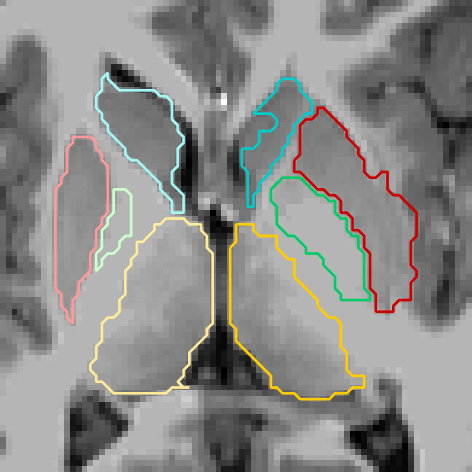}}
            \shortstack{Our CNN \\
                \includegraphics[width=0.35\textwidth]{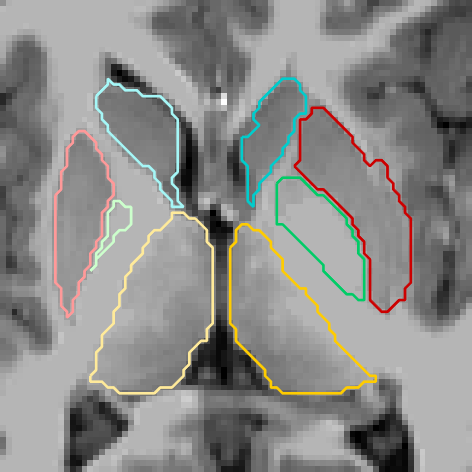}}
          
        }\\ 
        \mbox{
            \includegraphics[width=0.35\textwidth]{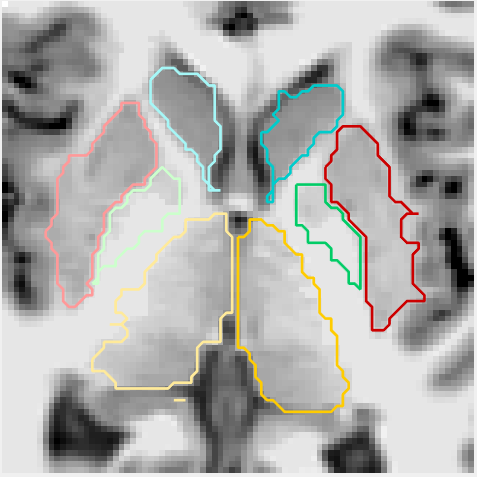}
            \includegraphics[width=0.35\textwidth]{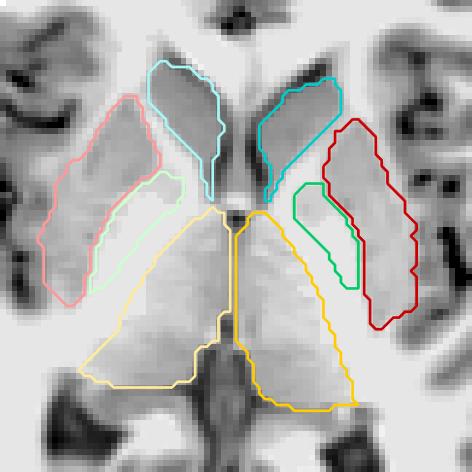}
           
        }\\ 
	    \mbox{
            \includegraphics[width=0.35\textwidth]{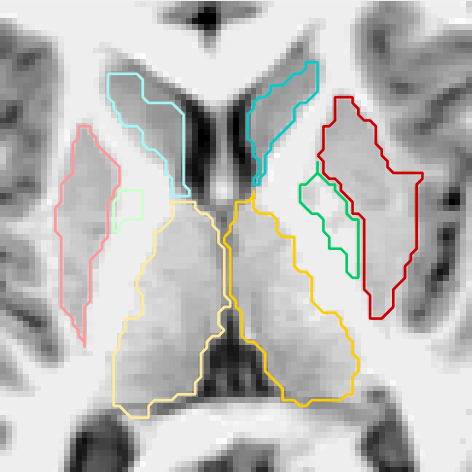}
            \includegraphics[width=0.35\textwidth]{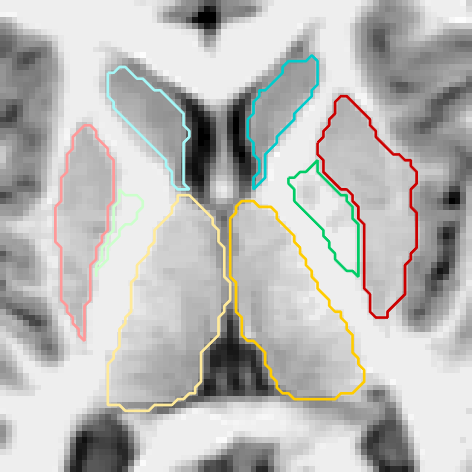}
        }
    \end{center}
    \caption{Visual examples of our 3D FCNN architecture compared with the references standard contoured by \FreeSurfer{} for three test subjects from sites not used in training.}%
   \label{fig:subfiguresSecondDataset}
\end{figure}

Figures \ref{fig:subfiguresFirstDataset} and \ref{fig:subfiguresSecondDataset} give visual examples of segmentations obtained by our 3D FCNN architecture and standard references contoured by \FreeSurfer{}.
 
It can be observed that the segmentations generated by our proposed architecture are significantly smoother than those of \FreeSurfer{}, regardless of the subject group (i.e diagnosis, age, site employed or not in training). We also notice that our system is better at identifying thin regions in the structures of interest, for instance, the lower extremities of pallidum (green regions).

\begin{figure}[ht!]
     \begin{center}
        \subfigure{%
        \label{fig:featMapsRight}
            \includegraphics[width=0.315\textwidth]{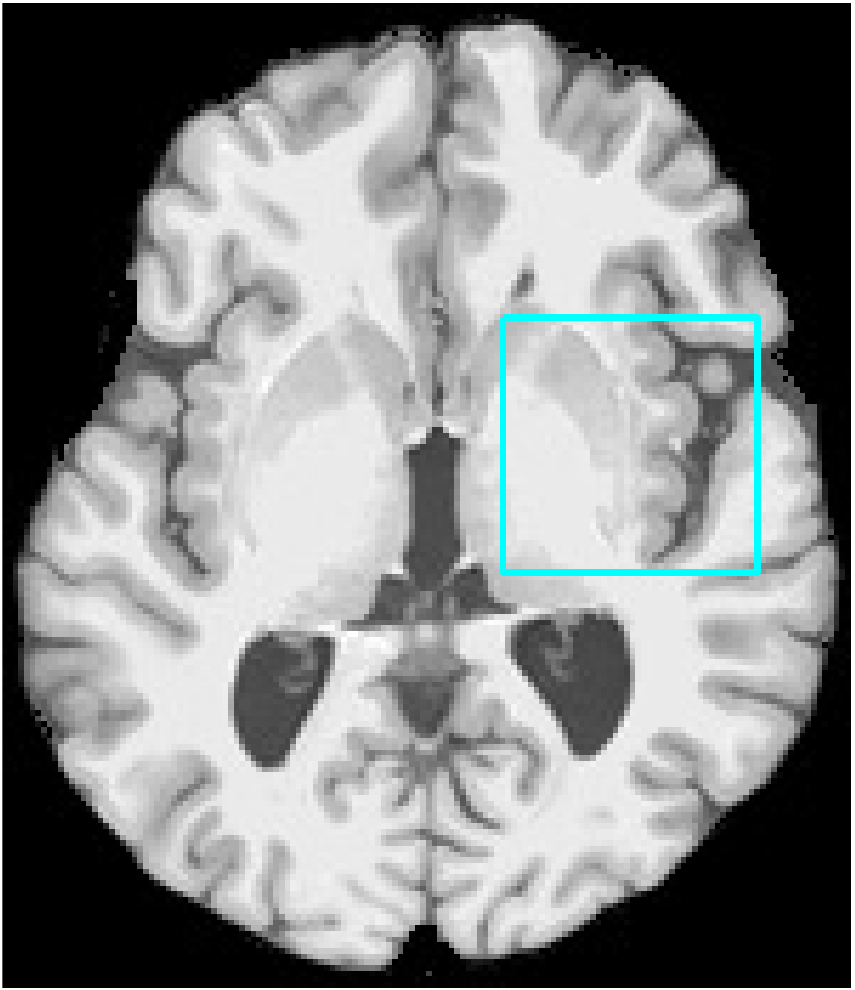}
        }%
        \hspace{-0.25em}
        \subfigure{%
         \label{fig:featMapsLeft}
            \includegraphics[width=0.642\textwidth]{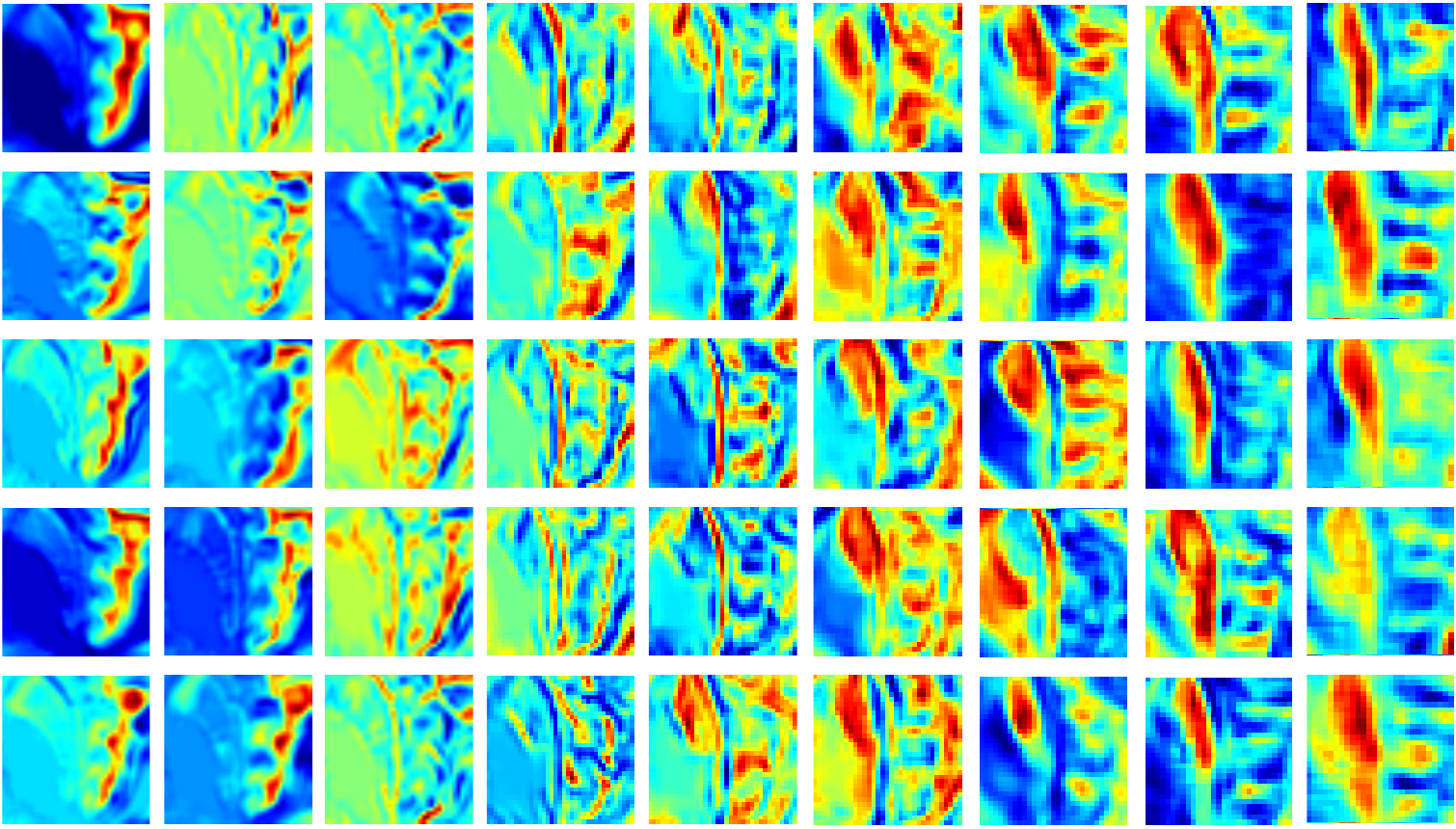}
        }%
         \end{center}
    \caption{Feature map activations in all convolutional layers of the FCNN (\emph{right}), obtained for a given patch of the input MRI image (\emph{left}). Each column corresponds to a different convolutional layer, from shallow to deeper, and each image in a row to a features map activation randomly selected in the layer.}%
   \label{fig:featMaps}
\end{figure}

To better understand the features learned by the network, Figure \ref{fig:featMaps} shows examples of feature map activations obtained for a given input patch (cyan box in the figure). Each column corresponds to a different CNN layer, left-side columns corresponding to shallow layers, and right-side columns to deep layers in the network. Likewise, images in each row correspond to a randomly selected activation of the layer's feature map. Although difficult to analyze, we notice that activation values in initial layers mainly indicate the presence of strong edges or boundaries, whereas those in deeper layers of the network represent more complex structures. In particular, images in the last two columns (i.e., convolutional layers of the network) roughly delineate the right caudate. Note that 2D images are used here for visualization purposes and that both input patches and features map activations are actually in 3D.

\begin{figure}[ht!]
\begin{small}
     \begin{center}

        \mbox{
            \shortstack{\includegraphics[width=0.18\textwidth]{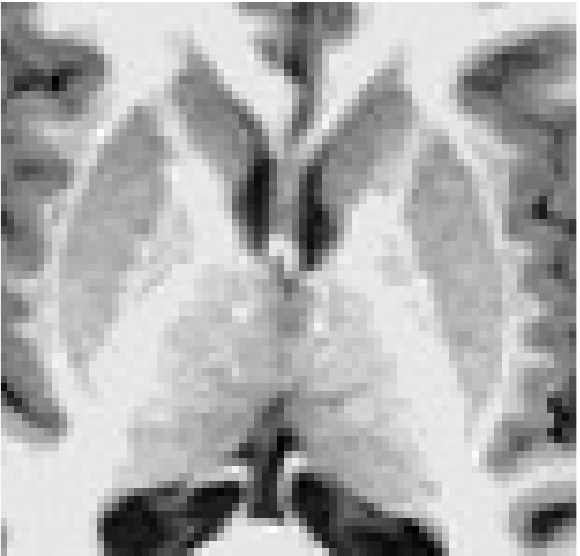} \\ MRI image}
            \shortstack{\includegraphics[width=0.18\textwidth]{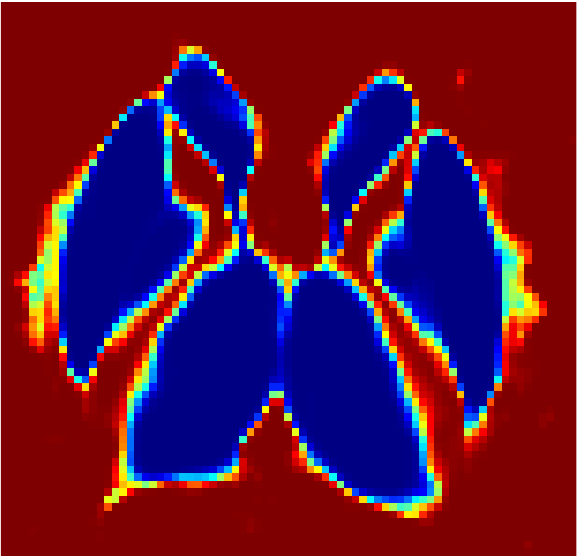} \\ Background}
            \shortstack{\includegraphics[width=0.18\textwidth]{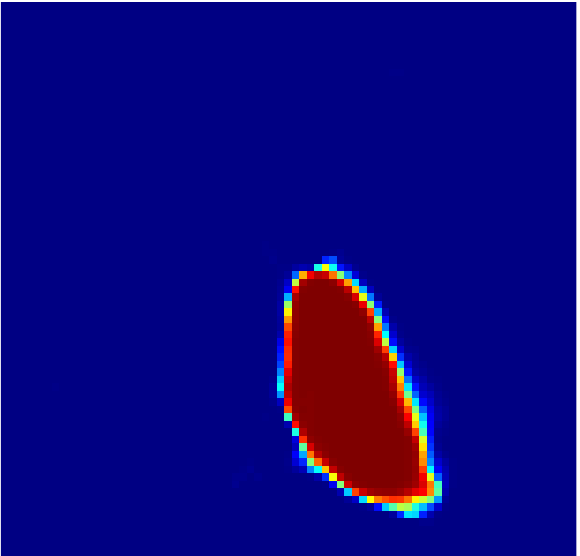} \\ Th(R)}
           \shortstack{\includegraphics[width=0.18\textwidth]{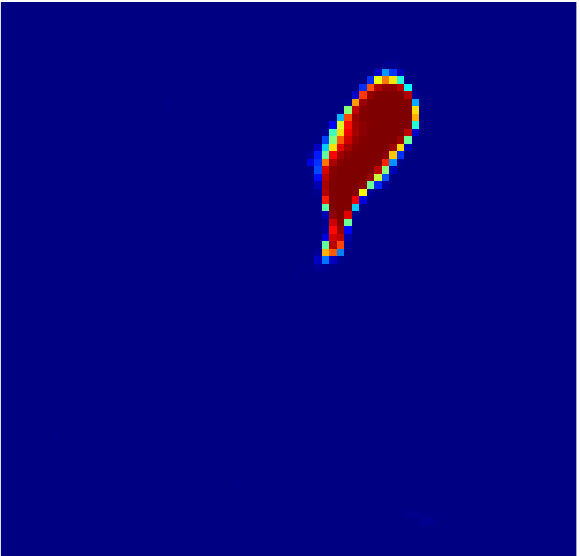} \\ Ca(R)}
           \shortstack{\includegraphics[width=0.18\textwidth]{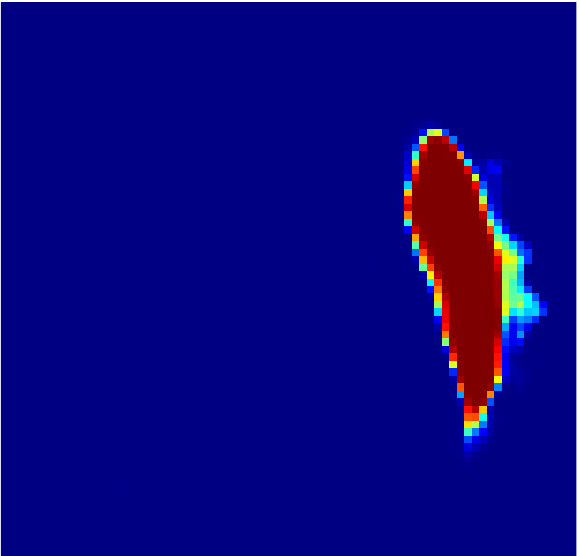} \\ Pu(R)}
        }\\ 
        
        \vspace{1mm}
        
        \mbox{
            \shortstack{\includegraphics[width=0.18\textwidth]{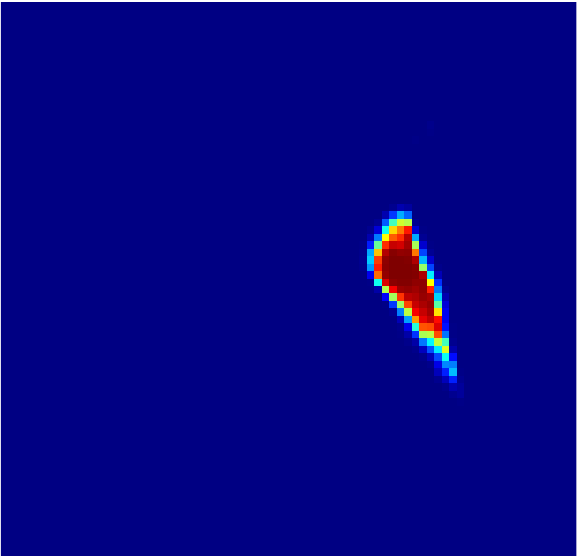} \\ Pa(R)}
            \shortstack{\includegraphics[width=0.18\textwidth]{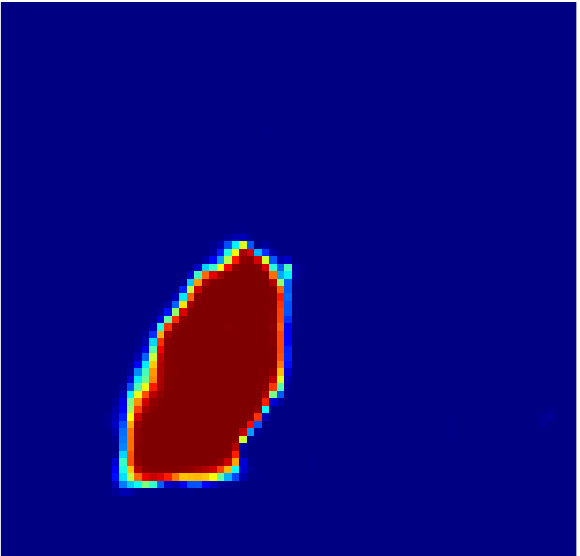} \\ Th(L)}
            \shortstack{\includegraphics[width=0.18\textwidth]{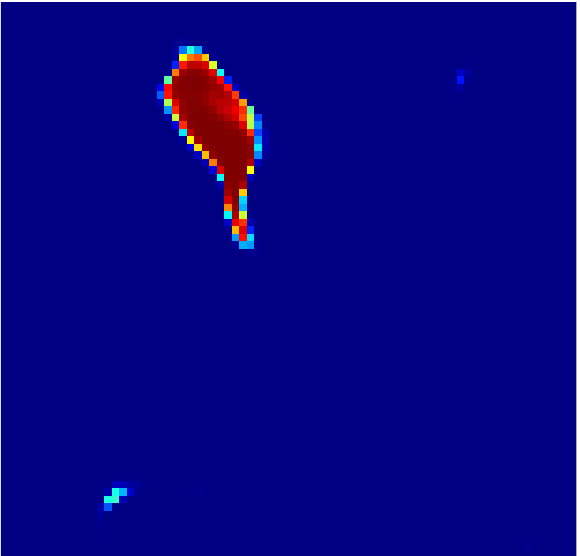} \\ Ca(L)}
            \shortstack{\includegraphics[width=0.18\textwidth]{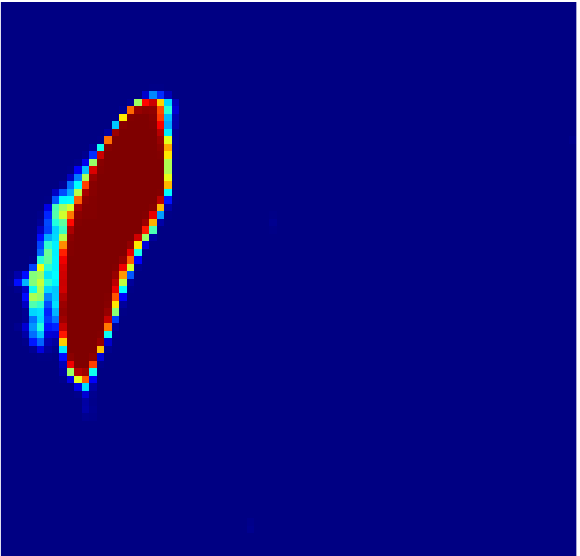} \\ Pu(L)}
            \shortstack{\includegraphics[width=0.18\textwidth]{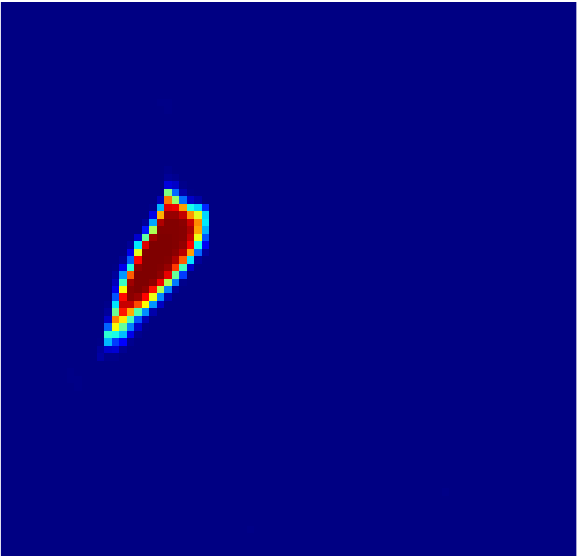} \\ Pa(L)}
        }\\
        
        \vspace{1mm}
        
        \mbox{
            \includegraphics[width=0.75\textwidth]{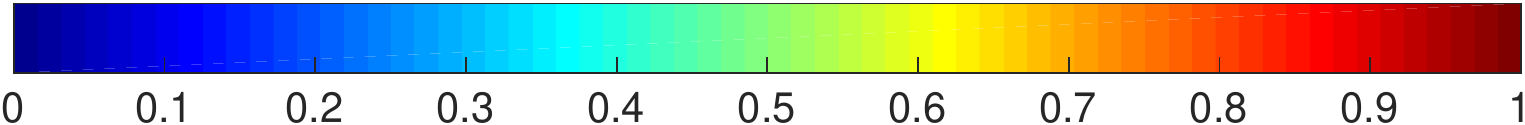}
        }
    \end{center}
    \end{small}    
    \caption{Probability maps generated by the proposed 3D FCNN for the background and the eight structures given an input MRI image. Note that input MRI has been cropped for better resolution.}%
   \label{fig:probMaps}
\end{figure}

\begin{figure}[ht!]
     \begin{center}
    
     \subfigure{%
        \includegraphics[height=3.5cm, width=3.5cm]{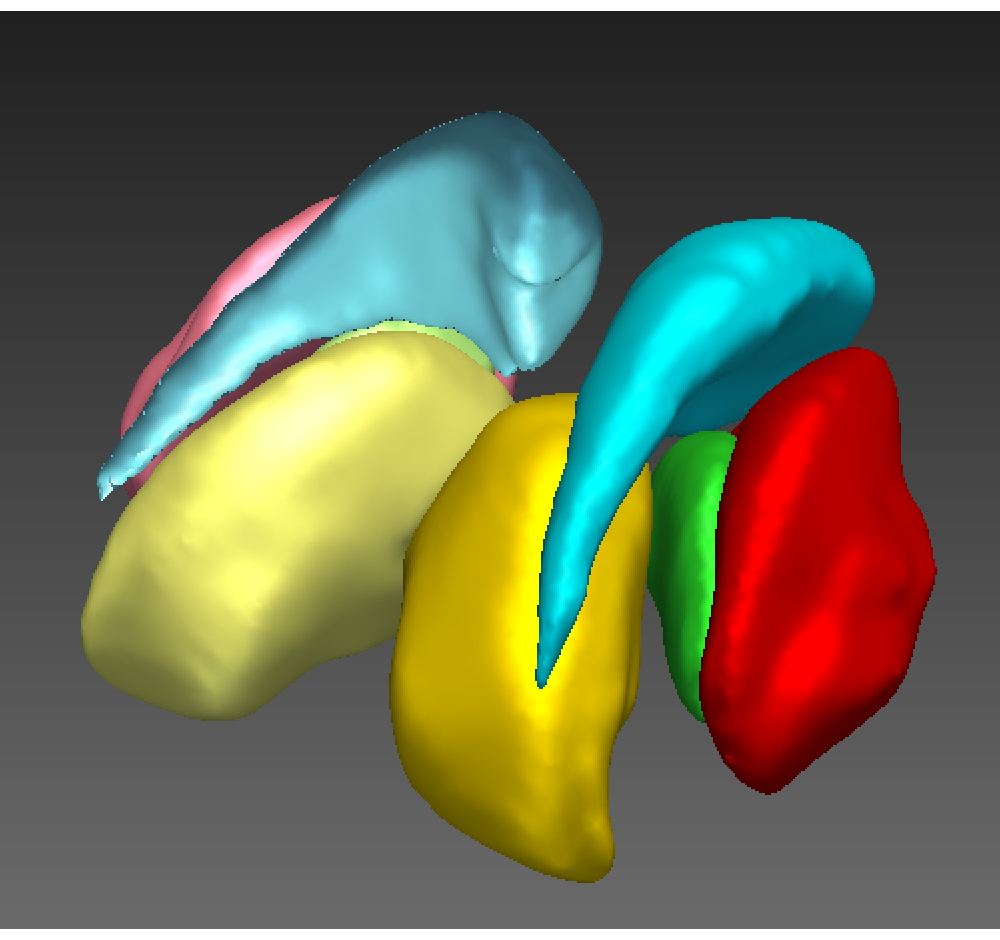}
        }
      \subfigure{%
        \includegraphics[height=3.5cm, width=3.5cm]{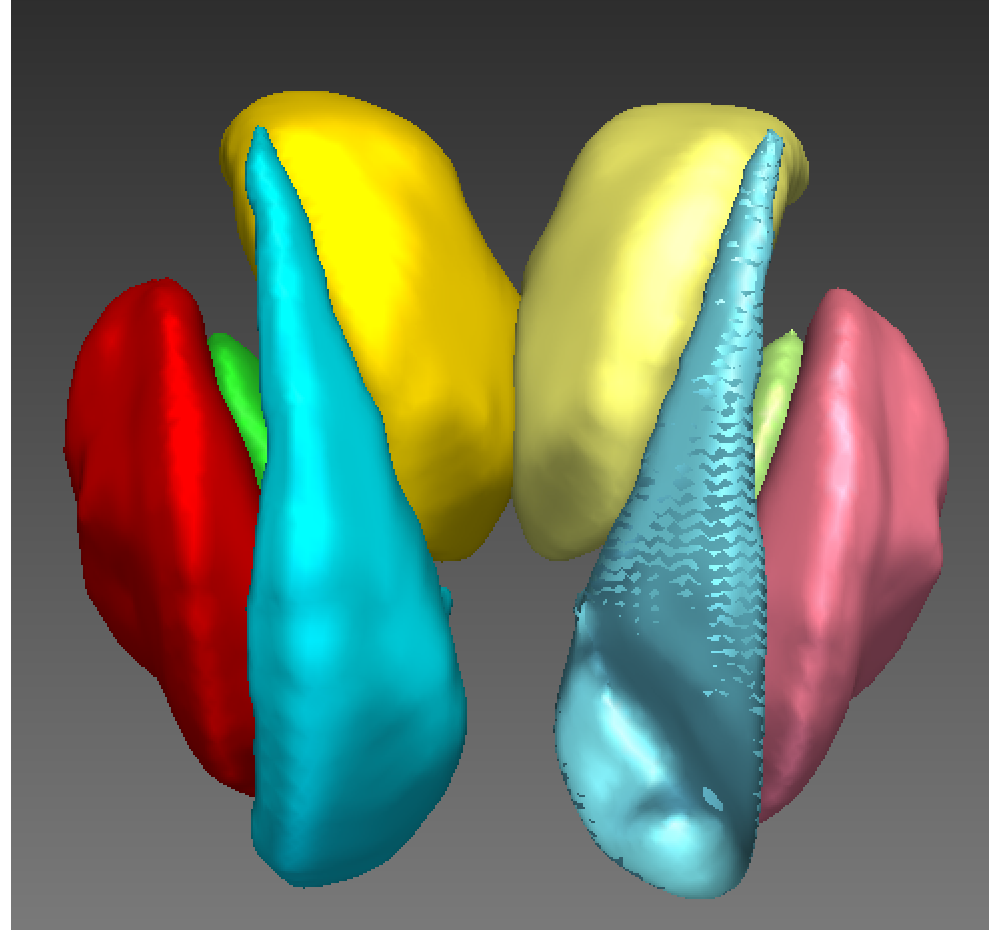}
        }
         \subfigure{%
        \includegraphics[height=3.5cm, width=3.5cm]{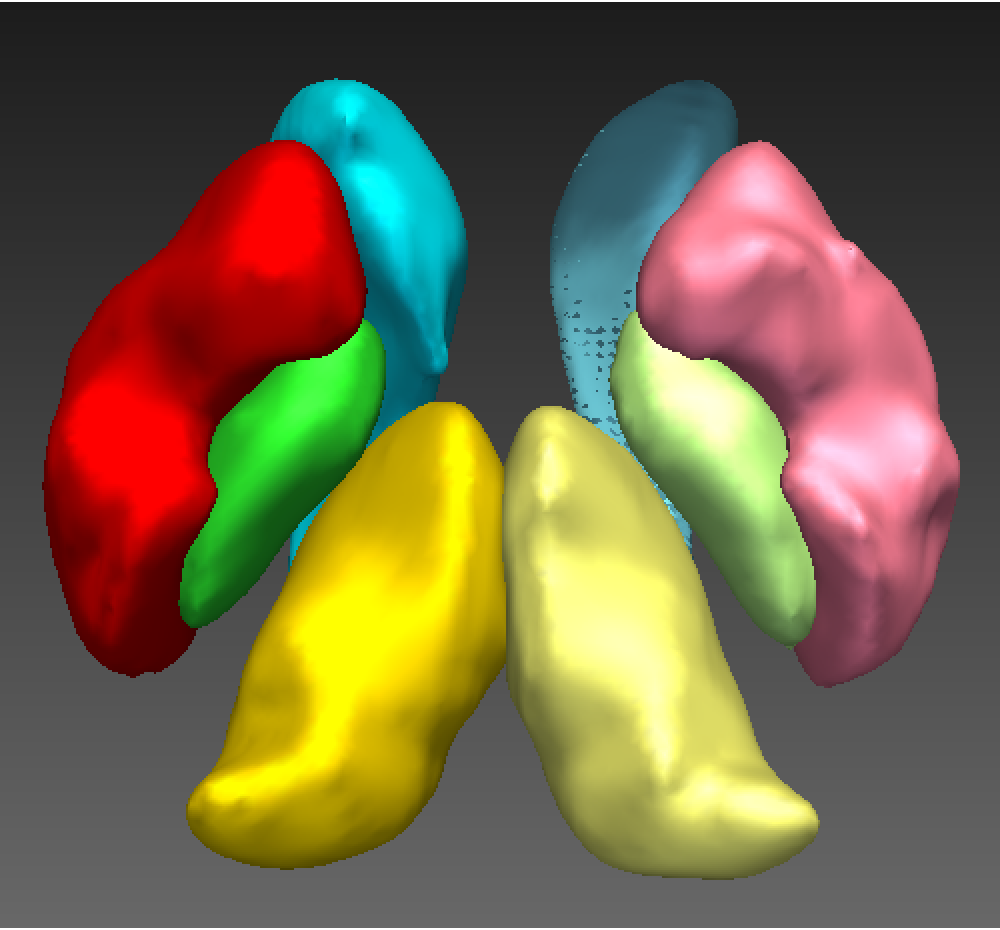}
        }
        \caption{Different views of a smoothed version of contours provided by our automatic segmentation system. In these images, the thalamus, caudate, putamen and pallidum are respectively depicted in yellow, cyan, red and green.}
        \label{fig:3D}
\end{center}        
\end{figure}

As previously explained, score maps (i.e., class probabilities, ranging from 0 to 1) are obtained at the end of the network, before the voxels are assigned to the target labels. To illustrate this output, Figure \ref{fig:probMaps} shows an example of probability maps for a given slice of the volume. Red pixels indicate probability values close to 1, and blue pixels near 0. Each image of the figure gives the probability map of a specific structure of interest, including the background. It can be seen that generated probability maps are well defined, reflecting the actual contours of the imaged structures (first subfigure of the set). This suggests that these probability maps can be used directly for segmentation, without requiring additional, and potential computationally expensive, spatial regularization. Smoothed examples of 3D segmentation outputs are displayed in Figure \ref{fig:3D}. These images, which were rendered using the Medical Interaction ToolKit (MITK) software package \cite{wolf2005medical}, highlight the spatial consistency of the obtained segmentation. All automatic contours and probability maps generated by our network are publicly available at: \sloppy\burl{https://github.com/josedolz/3D-FCNN-BrainStruct}.

Using the GPU mentioned in Section \ref{sssec:architecture}, our method takes on average 2-3 minutes to segment the data of a single subject from the ABIDE dataset (nearly two days for all 947 subjects). For a fair comparison with a CPU-based \FreeSurfer{} implementation, we evaluated our method's processing time when running on the CPU only. A CPU version of our 3D CNN required less than 5 minutes per subject, in all cases. These CPU times are much lower than the several hours required by
\FreeSurfer{}'s full segmentation pipeline \cite{khan2008freesurfer,huo2016consistent}. Note that recent \FreeSurfer{} versions provide GPU support for some steps of the segmentation pipeline. For instance, in a benchmark test, performing a linear volumetric registration (i.e., \textit{mri\_em\_register}) takes 3 minutes on a GPU, compared to 33 minutes on a CPU\footnote{\url{http://surfer.nmr.mgh.harvard.edu/fswiki/CUDADevelopersGuide}}. However, this step is much less expensive computationally than the deformable registration step (i.e., \textit{mri\_ca\_register}), which typically takes an order of magnitude longer than linear registration\footnote{\url{http://surfer.nmr.mgh.harvard.edu/fswiki/ReconAllRunTimes}}.

\subsection{Cross-dataset evaluation}
\label{ssec:IBSR_ABIDE}

To further assess the potential of the proposed method on new data, we used the model trained on the ABIDE dataset for segmenting volumes from the IBSR dataset. Results of this experiment, presented in Table \ref{table_IBSR_ABIDE}, suggest that our method is reliable when tested on a different dataset, with a mean DSC ranging from 0.79 (in the pallidum) to 0.87 (in the thalamus). Compared to both training and testing on the IBSR dataset, these results correspond to an average decrease of 0.05 in DSC and an average increase of 0.12 mm in MHD. A possible reason for this performance drop is the difference in voxel resolutions between the two datasets, particularly in terms of thickness. While thickness in images acquired across the various sites of the ABIDE project is in the range between 1.0 to 1.2 mm (with only two sites providing images with thickness outside this range, i.e., 0.86 and 1.3 mm), all the scans in IBSR had a thickness of 1.5 mm.

\begin{table}[ht!]
\begin{footnotesize}
\centering
\renewcommand{\arraystretch}{1.2}
\begin{tabular}{cccccc}
\toprule
 & \textbf{Training} & \textbf{Thalamus} & \textbf{Caudate} & \textbf{Putamen} & \textbf{Pallidum} \\ \hline
\multirow{2}{*}{Mean DSC} & IBSR &  0.92  & 0.91 & 0.90 &  0.83\\
& ABIDE  &  0.87  &   0.84    &  0.85   &   0.79 \\ 
\midrule
\multirow{2}{*}{Mean MHD}  & IBSR  & 0.13  &  0.21  & 0.18  & 0.26 \\
& ABIDE  &  0.25  & 0.32  & 0.34  &  0.38 \\
\bottomrule
\end{tabular}
\caption{Segmentation accuracy (mean DSC and MHD) obtained on the IBSR dataset by the proposed 3D FCNN model trained with the ABIDE dataset or with the IBSR dataset.}
\label{table_IBSR_ABIDE}
\end{footnotesize}
\end{table}

\section{Discussion}
\label{sec:discussion}

We conducted a comprehensive quantitative evaluation of our method using the publicly available IBSR and ABIDE datasets. The segmentation accuracy of our method was measured with respect to the ground-truth segmentations of the IBSR dataset, and compared to recently proposed methods for the task of brain parcellation. As reported in Table 5, our method obtained state-of-the-art performance, with mean DSC values ranging from 0.83 to 0.91 and mean MHD values between 0.13 mm and 0.26 mm (Figure 3). The ABIDE dataset was then used to demonstrate our method's reliability for large-scale datasets acquired at multiple sites, and measure the impact of various factors, including age, diagnosis group (i.e., healthy control or ASD). Considering all test subjects together, our method obtained segmentations consistent with those of \FreeSurfer{}, with mean DSC between 0.86 and 0.92 and mean MHD ranging from 0.14 mm to 0.22 mm, across the target brain structures. The accuracy of our $\CNNmulti$ architecture was statistically higher than two other tester architectures, which do not use multiscale features and small kernels (Table \ref{table:ABIDEThreeConf}).

Considering the diagnosis group of subjects, segmentations obtained for both control and ASD subjects were of high quality, with similar mean DSC and MHD values (Figure \ref{fig:subfiguresSecondDataset}). Since ASD subjects are likely to have morphological (e.g., volumetric) differences in brain regions like the putamen \cite{sato2013increased}, hippocampus \cite{nicolson2006detection} or amygdala \cite{schumann2004amygdala}, compared to healthy sujects, this suggests that our method is robust to such differences. Analyzing the results according to subject age group, we noticed a slightly lower segmentation accuracy for younger subjects. This is consistent the fact that the brain is continuously developing until adulthood, and that young subjects have a larger variability during their development process. However, it has been found that brain development in autism follows an abnormal pattern, with accelerated growth in early life, which results in brain enlargement during childhood \cite{aylward2002effects}. Therefore, there may be some intermediate states of brain development in early ages of control and ASD subjects that were not fully captured by the network during training. {blue}{Another reason to explain such differences is that template-based segmentation may not be flexible enough to adapt to these pathological differences.} 
Finally, by achieving a comparable performance on subjects from sites used in training and subject from other sites, we demonstrated that our method is robust to the various imaging parameters and protocols.

The automated segmentation of brain regions in MRI is a challenging task due to the structural variability across individuals. To tackle this problem, a broad range of approaches have been proposed during the last decade (Table \ref{tab:sumState}), many of which are based on atlases. Although atlas-based segmentation has been used successfully for subcortical brain structure segmentation, a single atlas is often unsuitable for capturing the full structural variability of subjects in a given neuroimaging study. Several strategies have been presented to overcome the limitation of single atlas segmentation, for instance using multiple atlases alongside label fusion techniques \cite{khan2011optimal}. Nevertheless, one of the main drawbacks shared by all atlas-based methods is their dependency to the image registration step, which is both time-consuming and prone to errors. Recent studies have reported segmentation times of up to several hours per subject when employing \FreeSurfer{} \cite{khan2008freesurfer,huo2016consistent}. In \cite{powell2008registration}, Powell et al. presented an approach based on artificial neural networks as an alternative to atlas-based methods. However, registration was also a key component of their segmentation scheme, thus having the same drawbacks as atlas-based techniques. Also using machine learning, a 2D FCNN was proposed in \cite{shakeri2016sub} for the task of subcortical brain parcellation. Although the registration of subjects volumes was not initially required, the authors tested their CNN on data pre-registered to the Talairach space. As demonstrated by our experiments, our approach has the advantage of being alignment independent, a property of great importance when working with multi-subject or multi-site data. 

\begin{table}[h!]
\centering
\tiny
\rowcolors{1}{gray!25}{white}
\begin{tabular} {lllll}
\specialrule{1pt}{0pt}{0pt}
\rowcolor{gray!50}

\textbf{Work}   & \textbf{Method}   &  \textbf{Structures} &  \textbf{DSC} &  \textbf{Dataset}  \\
\specialrule{0.5pt}{0pt}{0pt}
\addlinespace
\parbox{2.7cm}{\textbf{Heckemann et al. \cite{heckemann2006automatic} \\ (2006) }} & 
\parbox{1.5cm}{Atlas} &
\parbox{1.5cm}{Thalamus \\ Caudate \\ Putamen \\ Pallidum  }   &
\parbox{1.0cm}{ 0.90\\0.90\\0.90\\0.80}   &
\parbox{2cm}{Own dataset}\\
\addlinespace
\parbox{2.7cm}{\textbf{Han et al. \cite{han2007atlas} \\ (2007) }} & 
\parbox{1.5cm}{Atlas} &
\parbox{1.5cm}{Thalamus \\ Caudate \\ Putamen \\ Pallidum  }   &
\parbox{1.0cm}{ 0.88\\0.84\\0.85\\0.76}   &
\parbox{2cm}{Own dataset}\\
\addlinespace
\parbox{2.7cm}{\textbf{Linguraru et al. \cite{linguraru2007segmentation} \\ (2007) }} & 
\parbox{1.5cm}{Atlas} &
\parbox{1.5cm}{Thalamus \\ Caudate \\ Putamen \\ Pallidum  }   &
\parbox{1.0cm}{ 0.88\\0.82\\0.86\\0.79}   &
\parbox{2cm}{IBSR}\\
\addlinespace
\parbox{2.7cm}{\textbf{Bazin et al. \cite{bazin2008homeomorphic} \\ (2008) }} & 
\parbox{1.5cm}{Atlas} &
\parbox{1.5cm}{Thalamus \\ Caudate \\ Putamen \\ Pallidum  }   &
\parbox{1.0cm}{ 0.77\\0.78\\0.82\\-}   &
\parbox{2cm}{IBSR}\\
\addlinespace
\parbox{3cm}{\textbf{Powell et al. \cite{powell2008registration} \\ (2008)}} & 
\parbox{1.5cm}{Artificial \\ Neural \\ Network} &
\parbox{1.5cm}{Thalamus \\ Caudate \\ Putamen \\ Pallidum  }   &
\parbox{1.0cm}{ 0.88\\0.84\\0.85\\-}   &
\parbox{2cm}{Own dataset}\\
\addlinespace
\parbox{3cm}{\textbf{Artaechevarria et al. \cite{artaechevarria2009combination} \\ (2009)}} & 
\parbox{1.5cm}{Atlas} &
\parbox{1.5cm}{Thalamus \\ Caudate \\ Putamen \\ Pallidum  }   &
\parbox{1.0cm}{ 0.88\\0.83\\0.87\\0.81}   &
\parbox{2cm}{IBSR}\\
\addlinespace
\parbox{3cm}{\textbf{Ciofolo et al. \cite{ciofolo2009atlas} \\ (2009)}} & 
\parbox{1.5cm}{Atlas} &
\parbox{1.5cm}{Thalamus \\ Caudate \\ Putamen \\ Pallidum  }   &
\parbox{1.0cm}{ 0.77\\0.60\\0.66\\0.56}   &
\parbox{2cm}{IBSR}\\
\addlinespace
\parbox{3cm}{\textbf{Lotjonen et al. \cite{lotjonen2010fast} \\ (2010)}} & 
\parbox{1.5cm}{Atlas} &
\parbox{1.5cm}{Thalamus \\ Caudate \\ Putamen \\ Pallidum  }   &
\parbox{1.0cm}{ 0.89\\0.85\\0.90\\0.80}   &
\parbox{2cm}{IBSR}\\
\addlinespace
\parbox{3cm}{\textbf{Sabuncu et al. \cite{sabuncu2010generative} \\ (2010)}} & 
\parbox{1.5cm}{Atlas} &
\parbox{1.5cm}{Thalamus \\ Caudate \\ Putamen \\ Pallidum  }   &
\parbox{1.0cm}{ 0.91\\0.87\\0.89\\0.84}   &
\parbox{2cm}{Own dataset}\\
\addlinespace
\parbox{3cm}{\textbf{Patenaude et al. \cite{patenaude2011bayesian} \\ (2011)}} & 
\parbox{1.5cm}{Bayesian model} &
\parbox{1.5cm}{Thalamus \\ Caudate \\ Putamen \\ Pallidum  }   &
\parbox{1.0cm}{ 0.89\\0.83\\0.88\\0.79}   &
\parbox{2cm}{Own dataset}\\
\addlinespace
\parbox{3cm}{\textbf{Rousseau et al. \cite{rousseau2011supervised}\\ (2011) }} & 
\parbox{1.5cm}{Atlas} &
\parbox{1.5cm}{Thalamus \\ Caudate \\ Putamen \\ Pallidum  }   &
\parbox{1.0cm}{ 0.88\\0.87\\0.87\\0.64}   &
\parbox{2cm}{IBSR}\\
\addlinespace
\parbox{3cm}{\textbf{Asman et al. \cite{asman2014groupwise} \\ (2014)}} & 
\parbox{1.5cm}{Atlas} &
\parbox{1.5cm}{Thalamus \\ Caudate \\ Putamen \\ Pallidum  }   &
\parbox{1.0cm}{ 0.89\\0.90\\0.89\\0.84}   &
\parbox{2cm}{OASIS}\\
\addlinespace
\parbox{3cm}{\textbf{Wang et al. \cite{wang2014multi} \\ (2014) }} & 
\parbox{1.5cm}{Atlas} &
\parbox{1.5cm}{Thalamus \\ Caudate \\ Putamen \\ Pallidum  }   &
\parbox{1.0cm}{ 0.89\\0.75\\0.88\\0.84}   &
\parbox{2cm}{OASIS}\\
\addlinespace
\parbox{3cm}{\textbf{Shakeri et al. \cite{shakeri2016sub} \\ (2016)}} & 
\parbox{1.5cm}{2D FCNN \\ + CRF} &
\parbox{1.5cm}{Thalamus \\ Caudate \\ Putamen \\ Pallidum  }   &
\parbox{1.0cm}{ 0.87\\0.78\\0.83\\0.75}   &
\parbox{2cm}{IBSR}\\
\addlinespace
\addlinespace
\parbox{3cm}{\textbf{Bao et al. \cite{bao2016multi} \\ (2016)}} & 
\parbox{1.5cm}{2D CNN } &
\parbox{1.5cm}{Thalamus \\ Caudate \\ Putamen \\ Pallidum  }   &
\parbox{1.0cm}{ 0.90\\0.87\\0.88\\0.80}   &
\parbox{2cm}{IBSR}\\
\addlinespace
\parbox{3cm}{\textbf{Our CNN} } & 
\parbox{1.5cm}{3D FCNN } &
\parbox{1.5cm}{Thalamus \\ Caudate \\ Putamen \\ Pallidum  }   &
\parbox{1.0cm}{ \textbf{0.92}\\\textbf{0.91}\\\textbf{0.90}\\\textbf{0.83}}   &
\parbox{2cm}{IBSR}\\
\addlinespace
\parbox{3cm}{\textbf{Our CNN} } & 
\parbox{1.5cm}{3D FCNN } &
\parbox{1.5cm}{Thalamus \\ Caudate \\ Putamen \\ Pallidum  }   &
\parbox{1.0cm}{ \textbf{0.92}\\\textbf{0.92}\\\textbf{0.91}\\\textbf{0.86}}   &
\parbox{2cm}{ABIDE}\\
\addlinespace
\specialrule{1pt}{0pt}{0pt}
\end{tabular}
\caption{Summary of brain subcortical structures segmentation methods. While most of these methods employ IBSR for evaluation purposes, OASIS public dataset and proprietary datasets have been also employed.}
\label{tab:sumState}
\end{table}

Although 2D CNNs have led to record-breaking performances in various computer vision tasks, their usefulness for 3D medical images is more limited. Numerous strategies have been proposed to mitigate this, for instance, considering all three orthogonal planes \cite{Brebisson2015deep}, or using single slices with a regularization scheme (e.g., CRF) to impose volumetric homogeneity \cite{shakeri2016sub}. While these techniques have helped improving segmentation results, they lack the ability to capture the full spatial context of 3D images. By using 3D convolutions, our approach can better capture spatial context in volumetric data. This is reflected by a performance improvement with respect to typical 2D CNN models. Another noteworthy point is the ability of our method to successfully segment subjects from sites that were not employed during training. Differences in scanners or acquisition protocols, for instance, can introduce a significant bias on the appearance of images (e.g., alignment, contrast, etc.), and the heterogeneity of multi-site data has been a stumbling block for large-scale neuroimaging studies. As confirmed by our results, incorporating training samples from different sites, which cover a wider range of variability, allowed us to alleviate this problem.

For the experiments on the ABIDE dataset, the reference contours used for training our CNN were obtained with \FreeSurfer{}, which is considered as a standard approach to subcortical brain labelling. While expert-labelled contours would have provided a more reliable validation of our approach, it was found that the contours obtained by our method were consistent with those of \FreeSurfer{}. Furthermore, a visual inspection of the results revealed that our method's contours were, in most cases, more regular than those obtained by \FreeSurfer{}. This suggests our method to be a suitable alternative to \FreeSurfer{}'s parcellation pipeline. Nevertheless, an evaluation involving trained clinicians would be necessary to fully validate this assertion. 

An interesting finding that can be observed from the experiments, is the comparable performance of the proposed 3D FCNN on both IBSR and ABIDE datasets (See table \ref{tab:sumState}, \textit{last two rows}). It is important to note that while IBSR subjects are pre-aligned, ABIDE subjects are not. This suggests that pre-alignment does not have a significant influence on the performance of the proposed architecture. 

Another important fact is that the performance was not affected by dataset size (IBSR is small whereas ABIDE is very large). 
This suggests that, in our case, data augmentation may not be of substantial benefit. This is due to our use of sub-volumes, which yields a large number of samples for each subject.

\begin{figure}[ht!]
     \begin{center}
    
     \mbox{
        \includegraphics[height=5cm]{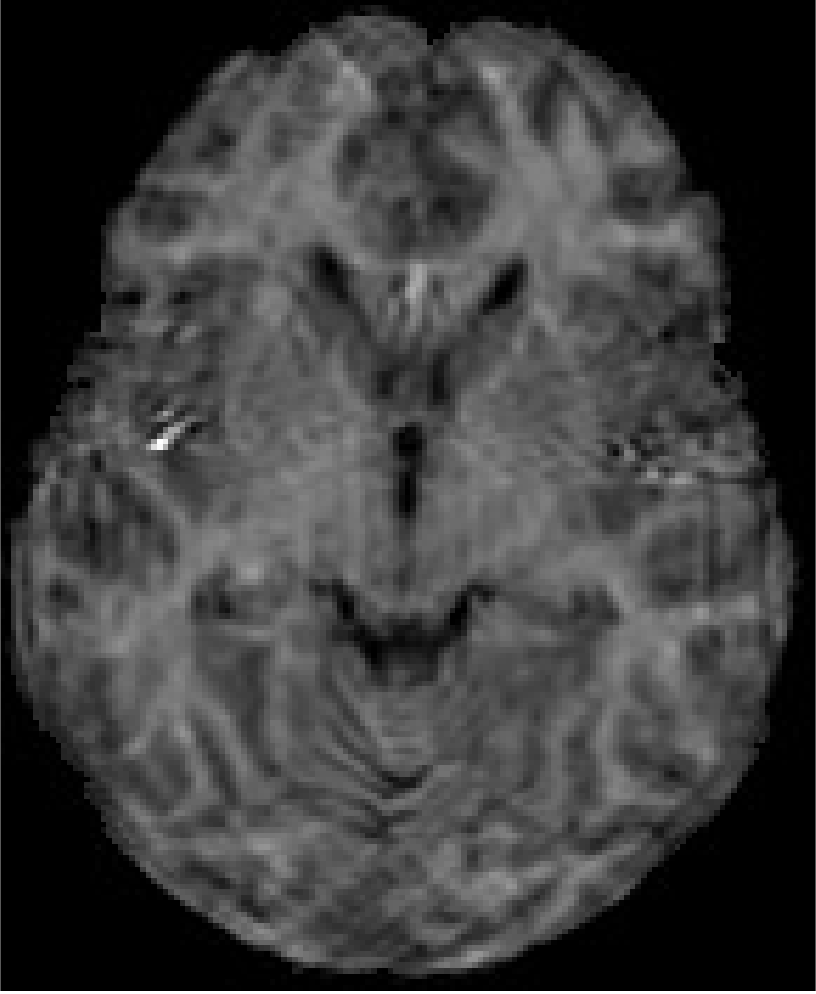}
        
        \hspace{2mm}
        
        \includegraphics[height=5cm]{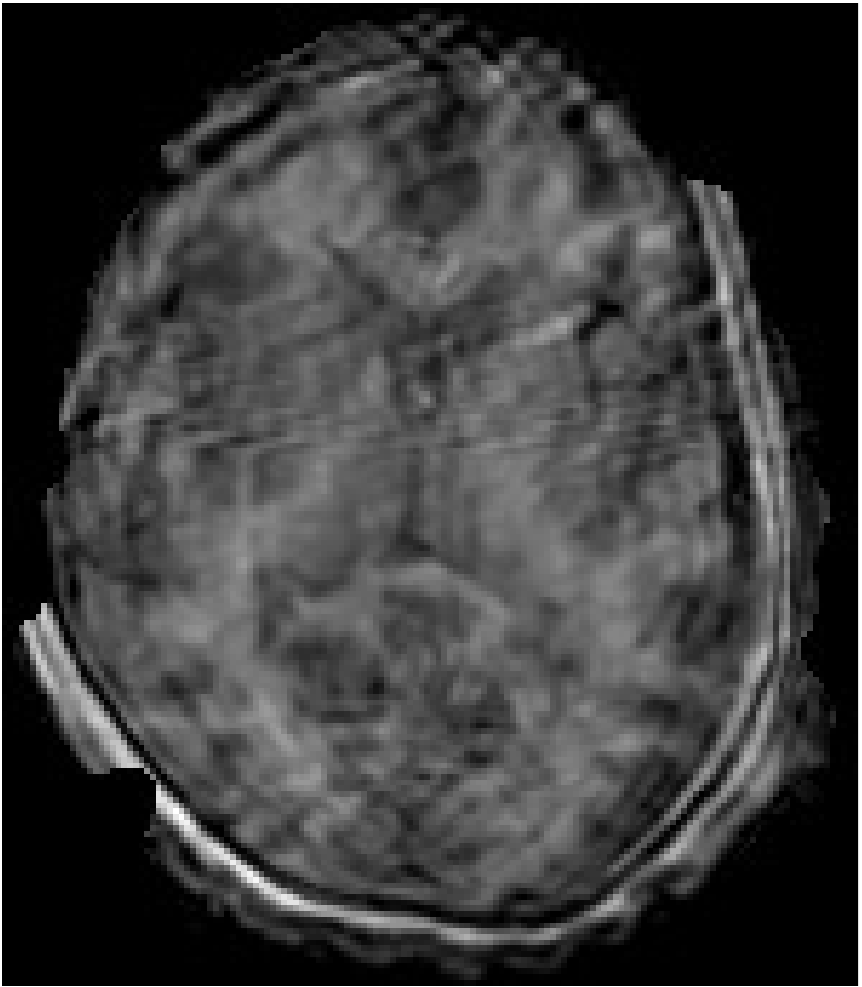}
        }
        \caption{Axial slices from a bad quality scan of two subjects that were excluded from the evaluation. }
        \label{fig:fail}
\end{center}        
\end{figure}

Analyzing the results, we observed that the segmentation of several subject data differed considerably from others. Upon visual inspection, we found that the corresponding MRI images had a poor quality (e.g., motion artifacts), and decided not to include them in the evaluation. Figure \ref{fig:fail} shows examples of 2D slices (in axial view) of two subjects with problematic data.   

Many modifications to the proposed architecture are possible. For example, the architecture could have a different number of convolutional/fully-connected layers, or a different number of filters/units in these layers. Several parameters settings were tested in preliminary experiments to come up with a definitive architecture. Although the chosen parameters were found to perform well on the test data, they might not be optimal for other datasets. Despite this, small variations in the architecture are unlikely to have a large impact on performance. In future work, it would be interesting to further investigate the optimization of these parameters, such that they could be tuned automatically for a specific task and target data. In \cite{kamnitsas2016efficient}, Kamnitsas et al. found that different segment sizes as input to their network led to differences in performance. In our study, we used input sizes that worked well for their specific application, i.e. brain lesion segmentation. Although our target problem also uses brain images, characteristics of both problems are different, and the effect of input sizes on performance might also differ. We thus intend to investigate the impact of this factor in a subsequent study. 

A notable limitation of our architecture is its small receptive field, constrained by the significant memory requirements of 3D convolutions. As discussed earlier, some approaches have overcome this limitation by employing additional path-ways with lower image resolutions as input. However, this strategy comes at the cost of losing fine-grained information in the segmentation. In future work, we plan to investigate the use of dilated convolutional kernels \cite{yu2015multi} to enlarge the receptive field without losing resolution or increasing the number of trainable parameters.

Another important aspect of CNNs is the transferability of knowledge embedded in the pre-trained architectures, i.e transfer learning. The use of pre-trained CNNs has been already investigated in previous works. Nevertheless, available pre-trained models mainly come from 2D convolutions and its use is often tailored to the same application. We believe that pre-trained CNNs can be successfully used for different applications sharing the same nature, even if their objectives differ. For instance, our 3D FCNN trained on subcortical brain structures may be employed as pre-trained network to segment cardiac images. 

\section{Conclusion}
\label{sec:conclusion}
 
We presented a method based on fully-convolutional networks (FCNNs) for the automatic segmentation of subcortical brain regions. Our approach is the first to use 3D convolutional filters for this task. Moreover, by exploiting small convolution kernels, we obtained a deeper network that has fewer parameters and, thus, is less prone to overfitting. Local and global context were also modelled by injecting the outputs of intermediate layers in the network's fully-connected layers, thereby encouraging consistency between features extracted at different scales, and embedding fine-grained information directly in the segmentation process. 

We showed our multiscale FCNN approach to obtain state-of-the-art performance on the well-known IBSR dataset. We then evaluated the impact of various factors, including acquisition site, age and diagnosis group, using 1112 unregistered subject datasets acquired from 17 different sites. This \emph{large-scale} evaluation indicated our method to be robust to these factors, achieving outstanding accuracy for all subjects groups. Additionally, these experiments have highlighted the computational advantages of our approach compared to atlas-based methods, by obtaining consistent segmentation results in less time. In summary, we believe this work to be an important step toward the adoption of automatic segmentation methods in large-scale neuroimaging studies. 

\section*{Acknowledgments}

This work is supported by the National Science and Engineering Research Council of Canada (NSERC), discovery grant program, and by the ETS Research Chair on Artificial Intelligence in Medical Imaging.

\section*{References}


\end{document}